\documentclass[preprint,twocolumn,5p,numbers]{elsarticle}

\usepackage{amssymb}
\usepackage{amsthm}
\usepackage{lipsum}
\usepackage{amsmath} 
\usepackage{xcolor}
\usepackage{algorithm}
\usepackage{algorithmic}
\usepackage{comment}
\usepackage{graphicx}
\usepackage{subcaption}
\usepackage{caption}
\usepackage{listings}

\usepackage{hyperref}

\AtBeginEnvironment{tcolorbox}{\small}
\usepackage[most]{tcolorbox}
\usepackage{float}

\usepackage{mathtools}
\usepackage{comment}
\usepackage{multirow}
\usepackage{xspace}
\usepackage{hhline}

\usepackage{tikz}

\definecolor{darkblue}{rgb}{0.15,0.15,0.55}
\definecolor{lightgrey}{rgb}{0.75,0.75,0.75}

\definecolor{darkblue}{rgb}{0.15,0.15,0.55}
\definecolor{lightgrey}{rgb}{0.75,0.75,0.75}

\providecommand{\codecomment}[1]{\textcolor{lightgrey}{\dotfill}\textcolor{darkblue}{//\,\textrm{#1}}}

\newcommand\lsb[1]{{\color{blue}{#1}}}

\newcommand{\nphard}{$\mathcal{NP}$-hard\xspace}

\newcommand{\thm}{\noindent \textbf{Theorem}\xspace}

\newcommand{\pf}{\noindent \textbf{Proof}\xspace}

\journal{Robotics and Autonomous Systems}

\begin{document}

\begin{frontmatter}

\title{Very Large-scale Multi-Robot Task Allocation in Challenging Environments via Robot Redistribution}


\author[1]{Seabin Lee}
\ead{123sebin@naver.com}

\author[1]{Joonyeol Sim}
\ead{antihero1290@gmail.com}

\author[1]{Changjoo Nam\corref{cor1}}
\ead{cjnam@sogang.ac.kr}

\affiliation[1]{organization={Dept. of Electronic Engineering, Sogang University},
            addressline={Baekbeom-ro, Mapo-gu}, 
            city={Seoul},
            postcode={04107}, 
            country={South Korea}}
            
\cortext[cor1]{Corresponding author}

\begin{abstract}
We consider the Multi-Robot Task Allocation (MRTA) problem that aims to optimize an assignment of multiple robots to multiple tasks in challenging environments which are with densely populated obstacles and narrow passages. In such environments, conventional methods optimizing the sum-of-cost are often ineffective because the conflicts between robots incur additional costs (e.g., collision avoidance, waiting). Also, an allocation that does not incorporate the actual robot paths could cause deadlocks, which significantly degrade the collective performance of the robots. 

We propose a scalable MRTA method that considers the paths of the robots to avoid collisions and deadlocks which result in a fast completion of all tasks (i.e., minimizing the \textit{makespan}). To incorporate robot paths into task allocation, the proposed method constructs a roadmap using a Generalized Voronoi Diagram. The method partitions the roadmap into several components to know how to redistribute robots to achieve all tasks with less conflicts between the robots. In the redistribution process, robots are transferred to their final destinations according to a push-pop mechanism with the first-in first-out principle. From the extensive experiments, we show that our method can handle instances with hundreds of robots in dense clutter while competitors are unable to compute a solution within a time limit.
\end{abstract}


\end{frontmatter}

\section{Introduction}

\begin{figure}[!h]
    \centering
    \includegraphics[width=0.42\textwidth]{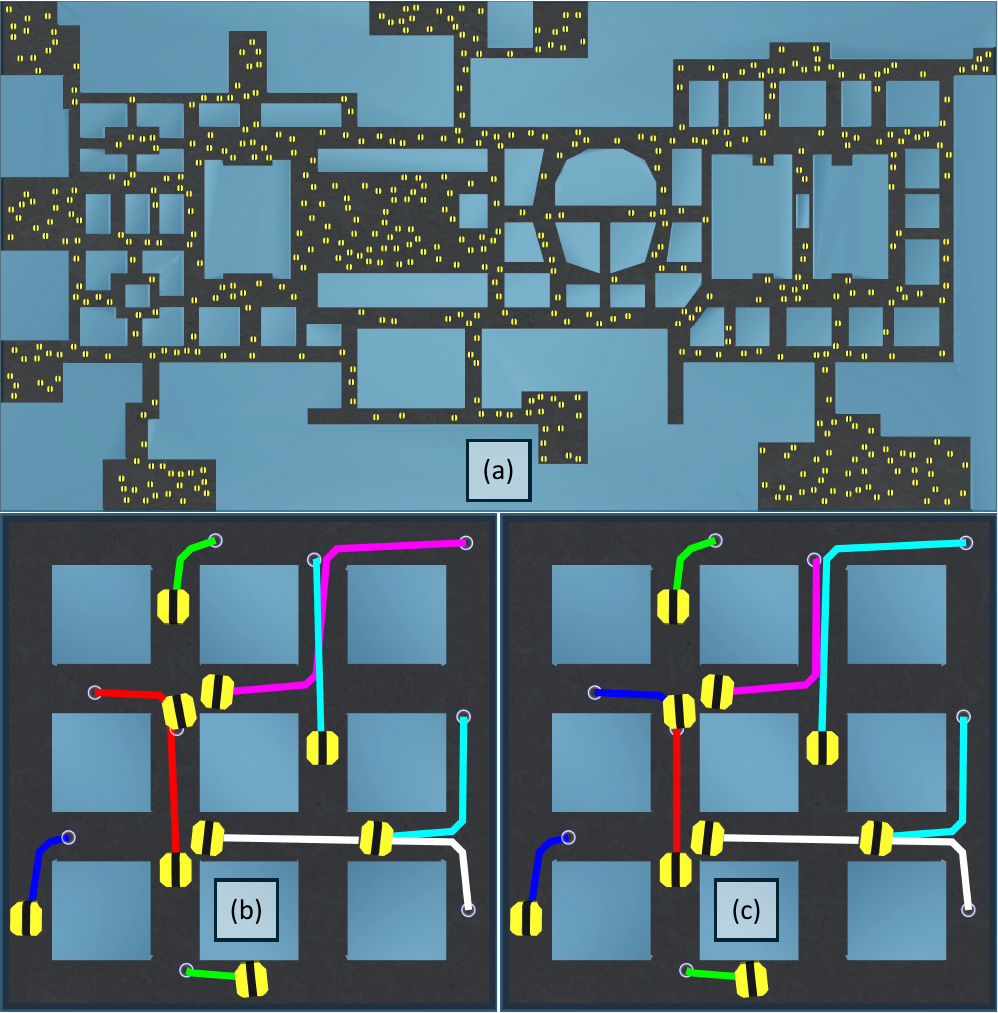}
    \caption{An example of challenging environments with the effect of considering inter-robot conflicts in the task allocation. (a) 500 robots in a shopping mall with narrow passages, which is one of our test instances incurring frequent conflicts. (b) An allocation found by a conventional method (the Hungarian method) that does not consider conflicts between robots. The colored lines represent expected paths from the robots to their assigned tasks that do not account for collisions between robots. A robot with a red line will have its path blocked by another robot on the path. (c) A result of the proposed method. The robots do not pass the corridors in the opposite direction, which is beneficial to prevent deadlocks.}
    \label{fig:page1}
    \vspace{-15pt}
\end{figure}

Multi-robot task allocation (MRTA) considers optimizing the collective performance of a team of robots that execute a set of tasks. In challenging environments with narrow passages and densely populated obstacles as illustrated in Fig.~\ref{fig:page1}a, an allocation of robots to tasks should consider the conflicts between robots when they navigate the environment. Otherwise, the robots may experience a long delay due to the conflicts or would face a deadlock situation which results in incomplete tasks as some robots could not proceed anymore.

The aforementioned problem can be approached in several ways depending on the optimization criteria. Assignments to minimizing the sum of costs can be found in polynomial time~\cite{gerkey2004formal}, however, minimizing the maximum cost (i.e., the minimax assignment problem) of performing tasks among all robots is shown to be \nphard~\cite{lenstra1990approximation}. While the canonical form of the minimax assignment problem is with costs that are constants, the MRTA problem in our interest could suffer from the costs that could vary even after an allocation is determined. Since the paths of robots to execute tasks are computed based on which tasks are allocated, the task costs directly related to the paths could not be exactly estimated unless we enumerate all cases with factorially many permutations. Thus, an optimal allocation computed using the cost estimates at planning time could be suboptimal at runtime.

As an effort to fill this optimality gap, Conflict-Based Search (CBS)~\cite{cbs} has been employed to solve the MRTA problem (CBS-TA)~\cite{cbs-ta} by incorporating path planning into MRTA so as to find a conflict-free allocation. Also, methods for multi-agent pickup and delivery (MAPD) problems (\cite{chen2021integrated}, \cite{liu2021integrated}) also consider the MRTA and the path planning problem together. However, their performance is limited to mid-sized instances with few tens of robots. Also, they require discretization of the workspace and action space of the robots so could oversimply the original problem in the continuous environment with physical robots. 

We propose a method called \underline{M}ulti-\underline{R}obot \underline{T}ask \underline{A}llocation via Robot \underline{R}edistribution \underline{M}echanism (MRTA-RM) that allocates robots to tasks while minimizing the conflicts between robots. MRTA-RM makes use of a roadmap to estimate the actual costs of robots to execute the assigned tasks. The method performs a redistribution of robots based on a demand-supply analysis to send robots to the part of the environment which lacks robots. The redistribution process eliminates the opposite-direction encounters of robots so that the robots can have less deadlocks achieving a shorter makespan (i.e., the completion time of all tasks). For example, the allocation found by a conventional method (e.g., Hungarian method~\cite{hungarian}) in Fig.~\ref{fig:page1}b shows encounters of robots from the opposite direction which incur delays or deadlocks. On the other hand, the allocation from our method (in Fig.~\ref{fig:page1}c) does not have robots that pass the corridors in the opposite direction which is beneficial to shorten the makespan.

Our contribution includes:
\begin{itemize}
    \item \textbf{Congestion-aware task allocation}: The proposed method considers the paths of the robots to perform tasks by constructing a roadmap, which is partitioned into several components. Robots are redistributed between the components to achieve all tasks such that the opposite-direction encounters of the robots are prevented. Thus, the robots can avoid conflicts and deadlocks while they move toward their tasks, resulting in a short makespan. 
    \item \textbf{Scalability}: The full task allocation process runs fast to handle hundreds of robots and tasks even in dense cluttered environments. 
    \item \textbf{Extensive experiments}: We run large sets of experiments in three different packed environments and two types of robot/task distributions. Total six other methods are compared to show the advantages of our proposed method.
\end{itemize}

\section{Related Work}
\label{related_work}


There have been few methods that consider task allocation and path planning to optimize the allocation based on the actual paths of robots performing tasks. 

Conflict-Based Search with Optimal Task Assignment (CBS-TA)~\cite{cbs-ta} produces an optimal solution by finding an allocation and the conflict-free paths to perform the allocated tasks together. In other words, it solves the Task Assignment and Pathfinding (TAPF) problem, which jointly addresses both MRTA and Multi-Agent Pathfinding (MAPF). CBS-TA operates within a search forest, with each root node representing a distinct task assignment. Upon encountering a conflict in a task allocation, CBS-TA introduces additional constraints to resolve the conflict which leads to the creation of new sub-nodes in the search tree. This high-level search strategy facilitates efficient exploration of the combined space of task assignments and paths. Enhanced CBS-TA (ECBS-TA) is also proposed in~\cite{cbs-ta} to improve the efficiency by introducing a focal search. While CBS-TA and ECBS-TA outperform other methods that solve MRTA and MAPF separately, its performance in larger-scale or more complex environments has not been extensively studied. Task Conflict-Based Search (TCBS)~\cite{tcbs} is another CBS-based algorithm that handles the TAPF problem. This method solves the problem by tree search with constraints, which considers both task assignments and collision-free paths for each agent simultaneously. However, TCBS is only tested with up to four robots and finds sub-optimal solutions.

Vitality-driven Genetic Task Allocation~\cite{zhang2019multi} uses a genetic algorithm to optimize task assignments and applies specific strategies to avoid conflicts. Scheduling conflicts are resolved by adjusting task order, while path conflicts are handled through a waiting strategy for shorter-path robots. Similarly, the Goal Assignment and Planning~\cite{turpin2014} solves the lexicographic bottleneck assignment problem~\cite{lexbap} on the graph. After the tasks are assigned, each robot is prioritized based on its path to the allocated task. Subsequently, time offsets are computed and applied to each robot to refine the paths and avoid conflicts between robots.

Another work~\cite{liu2021integrated} presents a method for TAPF problem in a large-scale autonomous robot network. It predicts the travel time of robots by incorporating the robot density in the predetermined sectors of the environment and the probability of delays. While the method scales with the number of robots so can be used for real-time applications, it must find the predetermined sectors of the environment which does not seem to be done autonomously. For example, each sector must have an entryway comprising exits and entrances to adjacent sectors. Also, every connecting passage must be able to accommodate at least two robots at the same time. 

The aforementioned methods (i.e., \cite{cbs-ta}, \cite{liu2021integrated}, \cite{tcbs}, \cite{zhang2019multi}) have limitations such that they only work in discrete space. However, our proposed method does not need hand-crafted structures of the environment but can find a solution in the continuous space. Instead of explicitly finding conflict-free paths, we find an allocation that is unlikely to have conflicts between robots by eliminating the opposite-direction robot encounters and the obstruction of moving robots. By doing so, our method can find a high-quality solution quickly even with hundreds of robots and tasks in environments with dense obstacles.

\section{Problem description}
\label{sec:description}



In this work, we consider navigation tasks of robots where assigning one robot to one task is sufficient to complete the task. The robots are interchangeable so any robot can accomplish any task. We also consider the MRTA problem which needs one-time assignment for a set of $M$ tasks and a team of $N$ robots.\footnote{Our method can be extended to find time-extended assignments by repeatedly running the proposed method until all tasks are performed.} We assume that the map and the locations of robots and tasks are known to the centralized agent computing the solution. We do not consider kinodynamics constraints of the robots. Also, no two or more tasks can be spatially overlapped. Likewise, robots do not start from the same location.

Our problem can be formulated as an assignment problem that aims to minimize the maximum cost incurred by performing any task by any robot, which is shown to be \nphard~\cite{lenstra1990approximation,chakrabarty20141}. The cost of performing task $t_j$ in task set $\mathcal{T}$ by robot $r_i$ in robot set $\mathcal{R}$ is denoted as $c_{ij} \in \mathbb{R}_{\ge 0}$ where $i \in \{1, \cdots, N\}$ and $j \in \{1, \cdots, M\}$. In our work, a cost means the actual time to complete a task by a robot at runtime, which is not an estimate at planning time. Therefore, the problem is to find an assignment of robots to tasks while minimizing the makespan (i.e., the elapsed time to complete all tasks). In the robotics problem with navigation tasks, this minimax assignment problem is even more difficult as $c_{ij}$ could vary depending on the paths taken by the robots to perform the assigned tasks.

We introduce a decision variable $x_{ij} \in \{0, 1\}$ which indicates an allocation of $r_i$ to $t_j$ if $x_{ij} = 1$. Without loss of generality, we assume that $N=M$ for simplicity (dummy robots or tasks can be added if $N \neq M$ if they are not balanced). Given this notation, the MRTA problem with the allocation result $\mathcal{A}$ can be defined as
\begin{equation} \label{eqn:def}
\centering
\begin{aligned}
\mbox{minimize} \quad \max\{c_{ij} x_{ij}\}, \qquad \forall \{i,j\}
\end{aligned}
\end{equation}
\noindent subject to
\begin{align} 
\sum_{j=1}^{M} x_{ij} &= 1, && \forall i,  \label{eqn:constraints1}\\[4pt]
\sum_{i=1}^{N} x_{ij} &= 1, && \forall j, \label{eqn:constraints2}\\[4pt]
x_{ij} &\in \{0, 1\}, && \forall \{i, j\}. \label{eqn:integerconstraint}
\end{align}

Given a workspace $\mathcal{W}$, we have a set of free space $\mathcal{W}_f = \mathcal{W} \setminus \mathcal{O}$ where $\mathcal{O}$ represents the set of obstacles. In $\mathcal{W}_f$, we define a topological graph $G=(\mathcal{V}, \mathcal{E})$ which we call a \textit{roadmap} where node $v \in \mathcal{V}$ is in $\mathcal{W}_f$ and $e \in \mathcal{E}$ connects a pair of nodes in $\mathcal{V}$. Since $G$ is a topological graph, every node of $G$ corresponds to a point in $\mathcal{W}_f$ and each edge represents a path connecting two points in $\mathcal{W}_f$~\cite{topological-graph}. 


Therefore, the roadmap encodes the traversable space in $\mathcal{W}$. We aim to solve (\ref{eqn:def}--\ref{eqn:integerconstraint}) and find all paths of robots to their assigned tasks on the roadmap while minimizing congestion during navigation. To achieve this, we aim to satisfy two constraints during robot execution:

\begin{itemize}
\item \textbf{Avoiding edge conflicts during traversal:} When multiple robots navigate the same roadmap, it is crucial to prevent conflicting movements along the same edges in opposite directions. By ensuring unidirectional flow on edges, robots can move without collision risks. \label{obj1}
\item \textbf{Preventing obstruction at task destinations:} When robots arrive at their allocated tasks, they should not block the paths of other robots that are still en route to their targets. This guarantees smooth movement and avoids bottlenecks during execution. \label{obj2}
\end{itemize}


\section{Methods}
We propose a framework for task allocation including robot redistribution which offers distinct advantages. Compared to traditional methods that directly allocate robots to tasks, our method can mitigate inter-robot conflicts during the navigation of the robots at execution time. 

Given an environment, the proposed method begins with generating a roadmap (i.e., a topological graph) to represent the environment and partitioning the roadmap into several subgraphs called components. The method performs the demand-supply analysis to identify how many robots are insufficient or overflow in each component. Then the redistribution planning determines which robot moves to which component to balance the robots. The redistribution plan considers the routes the robots take for their transfer to prevent possible conflicts between the robots. Once the redistribution is completed, the tasks are allocated to the robots.

\paragraph{\textsc{Approach}}

To ensure deadlock-free task allocation, our MRTA-RM method leverages two key properties during roadmap-based execution:


We define the roadmap $G = (\mathcal{V}, \mathcal{E})$ as a topological graph where node $v \in \mathcal{V}$ represents a point in the traversable space and edge $e \in \mathcal{E}$ represents a path between two nodes. Given a robot $r_i$ assigned to a task $t_j$, its start location and destination correspond to two nodes on the roadmap, denoted as $(v_{\footnotesize \mbox{s}}$ and $v_{\footnotesize \mbox{d}}$, respectively, where $v_{\footnotesize \mbox{d}}$ is the node that matches the position of the assigned task $t_j$. 

The path for $r_i$ is represented as a sequence of nodes $\mathcal{P}_{i} = (v^i_{\text{s}}, v^i_1, v^i_2, \dots, v^i_{\text{d}})$, where each edge between consecutive nodes $(v^i_k, v^i_{k+1}) \in \mathcal{E}$ are path segment.

We ensure two main properties during this traversal:
\begin{itemize}
    \item Property 1: Suppose an edge $e$ in the graph $G$ connects two nodes $v_a$ and $v_b$, then for any two robots $r_i$ and $r_j$, if $r_i$ is traveling from $v_a$ to $v_b$ along edge $e$, $r_j$ never travels from $v_b$ to $v_a$ along the same edge $e$ at any time. Therefore, robots can avoid collisions on edges. \label{form:condition1}
    
    \item Property 2: Given $\mathcal{P}_i$ and $\mathcal{P}_j$, there could be $v^i_{\footnotesize \mbox{d}} = v^j_{\footnotesize \textit{k}}$ for an arbitrary $k$. In such case, $r_j$ should not occupy $v^j_{\footnotesize \textit{k}}$ and move along the following nodes in $\mathcal{P}_j$ when $r_i$ arrives at $v^j_{\footnotesize \textit{k}} = v^i_{\footnotesize \mbox{d}}$. Thus, the arrived robot does not obstruct other robots. \label{form:condition2}
\end{itemize}
\vspace{-5pt}

To validate these two properties, we provide formal proofs in Sec.~\ref{sec:prove}, where we demonstrate how our method inherently enforces these behaviors through structured task allocation.

\subsection{Environment partitioning}
\label{sec:partition}

\begin{figure}
    \centering
    \includegraphics[width=0.25\textwidth]{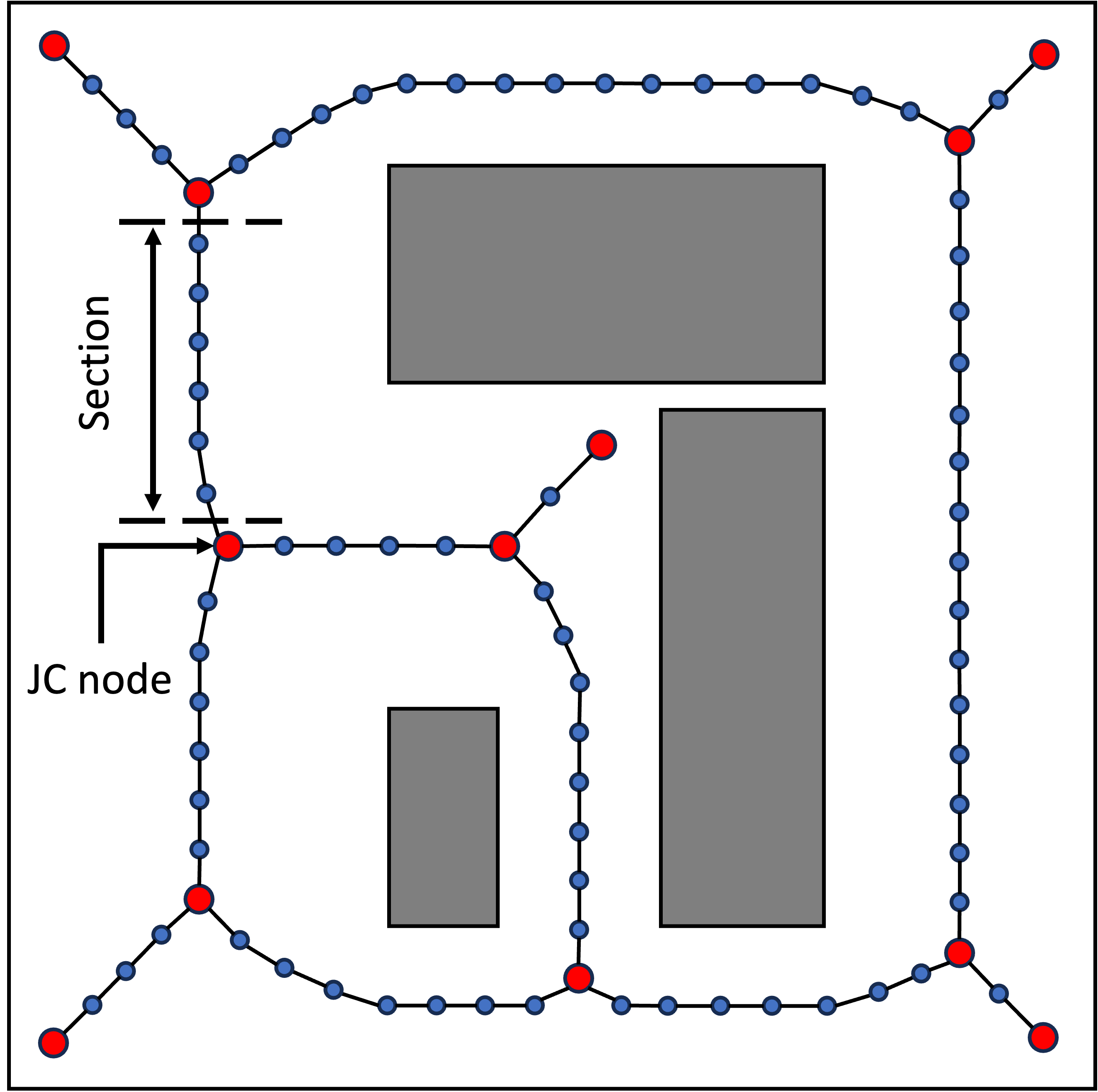}
    \caption{An example roadmap. Red larger nodes represent junction nodes (JC nodes). A section is defined by non-JC nodes between the JC nodes.}
    \label{fig:roadmap}
    \vspace{-17pt}
\end{figure}

Given a workspace $\mathcal{W}$, we construct a Generalized Voronoi Diagram (GVD). We employ a method proposed in~\cite{GVD} that uses sampled points from boundaries of obstacles to generate a GVD. The GVD is transformed into an initial topological graph which preserves the topological relationships between obstacles~\cite{topological-graph}. In line~\ref{line:JnS} of Alg.~\ref{alg:re-distribution}, a roadmap $G$ is generated after postprocessing such as removing nodes of the GVD that are not accessible to the robots (with a radius of $r$) owing to obstacles and distributing the nodes evenly, as shown in Fig.~\ref{fig:roadmap}. 

We then identify \textit{junction} nodes (JC nodes) and \textit{section} from the roadmap. The JC nodes (shown as red dots in Fig.~\ref{fig:roadmap}) are either nodes with more than two adjacent nodes or a terminal node (i.e., with one adjacent node). A section is uniquely defined by the nodes between two JC nodes, excluding the JC nodes themselves. Let $\mathcal{J}$ be the set of JC nodes, and $\mathcal{S}$ the set of all sections. The JC nodes and sections together form the \textit{components} of $G$, with $\mathcal{Z} = \mathcal{J} \cup \mathcal{S}$. The roadmap and its components remain the same unless new obstacles appear. Even if some obstacles occur, only affected components need to be updated.

\subsection{Demand and supply analysis}
\label{sec:demand}

For the demand and supply analysis between robots and tasks, we associated robots and tasks with the roadmap. Each robot belongs to exactly one node, which can be either a JC node or a non-JC node in some section. Likewise, a task is associated with a node. The association is determined by finding the closest node to the robot (or task), provided that the straight line connecting them does not intersect any obstacle in $\mathcal{O}$. This restriction ensures that a robot or task is not associated with a node that is spatially close but separated by an obstacle, such as a wall. Since the roadmap is constructed based on a GVD, which captures safe paths between obstacles, associating through obstacles would result in invalid or unreachable links in the roadmap.

In each section of the operational environment, nodes are arranged in a specific sequence according to their spatial positions. This ordered sequence begins with one of the end nodes, which is selected randomly to be the starting point, and is given the initial index of one. Subsequently, the adjacent node is assigned the next index, incrementing sequentially from the starting node. This process continues until all nodes within the section are indexed in a consecutive order.

Correspondingly, robots and tasks are indexed within their sections based on their spatial association with the nodes. The ordering of robots follows the node sequence. Robots (or tasks) associated with the first node in the section receive the earliest indices, and so on. In situations where multiple robots or tasks are linked to a single node, their ordering is refined based on their proximity to the node, with closer robots or tasks receiving lower indices, respectively. If multiple robots (or tasks) are located at the same distance to the node, the tie is broken by comparing their distances to the neighboring nodes connected to the associated node.Unlike sections, JC nodes do not provide a linear spatial context to determine relative positions among robots or tasks. Therefore, we index robots and tasks associated with JC nodes in the order they are encountered during roadmap traversal.


Then a component $z_s \in \mathcal{Z}$ (i.e., a JC node or a section in $G$) can be characterized by the numbers of robots and tasks belonging to $z_s$ where $s = 1, \cdots, |\mathcal{Z}|$. The quantity $D_s$ is the number of robots subtracted from the number of tasks associated with $z_s$. This value represents how many robots are overflowed or scarce for the component. Oversupplied component $z_s^+ \in \mathcal{Z}$ are with $D_s > 0$. Undersupplied component $z_s^- \in \mathcal{Z}$ are with $D_s < 0$. The rest of the components $z_s^\circ$ are balanced as $D_s = 0$. The set $\mathcal{Z}^+$ contains oversupplied components, and $\mathcal{Z}^-$ includes undersupplied components.

\subsection{Robot redistribution plan}
\label{sec:rm}

Once we finish the demand and supply analysis, we find a plan to redistribute the surplus robots in all $z_s^+$ to surplus tasks in all undersupplied components $z_s^-$. We employ a weighted matching algorithm computing an allocation $\mathcal{A}$ of the robots to the tasks\footnote{We use the Hungarian method~\cite{hungarian} but any alternative such as integer linear programming and auction algorithms should work.} to find the redistribution plan $\mathcal{X}$. The matching algorithm requires the costs of performing tasks by the robots. Often, many task allocation methods use simple cost measures such as the Euclidean distance between tasks and robots. However, the environments considered in this work are filled with obstacles. The actual paths for the robots to follow are seldom straight lines. Thus, we use the roadmap to estimate costs (i.e., travel distance) rather than using the Euclidean distances to reduce the optimality gap.

\begin{algorithm}[h]
\caption{{\scshape MRTA-RM}} \label{alg:re-distribution}
\begin{algorithmic}[1]
{\small
\floatname{algorithm}{Procedure}
\renewcommand{\algorithmicrequire}{\textbf{Input:} }
\renewcommand{\algorithmicensure}{\textbf{Output: }}
\REQUIRE $\mathcal{W}, \mathcal{O}, r, \mathcal{R}, \mathcal{T}$
\ENSURE Redistribution plan $\mathcal{X}$, Categories $\textsf{\small C1}, \textsf{\small C2}, \textsf{\small C3}, \textsf{\small C4}$
\STATE $\mathcal{P}=\emptyset$
\STATE $G, \mathcal{J}, \mathcal{S} \leftarrow$ \textsc{GenRoadmap}($\mathcal{W}, \mathcal{O}, r$)\codecomment{described in Sec.~\ref{sec:partition}} \label{line:JnS}
\STATE $\mathcal{Z} = \mathcal{J} \cup \mathcal{S}$
\STATE $\mathcal{Z}^+, \mathcal{Z}^- \leftarrow$ \textsc{Analysis}($G, \mathcal{O}, \mathcal{Z}, \mathcal{R}, \mathcal{T}$)\codecomment{described in Sec.~\ref{sec:demand}} \label{line:dns_analysis}
\FOR{each component $z_i^+ \in \mathcal{Z}^+$} \label{line:cost_start}
    \FOR{each component $z_j^- \in \mathcal{Z}^-$}
        \STATE $\mathcal{P}_{i,j} \leftarrow$ \textsc{ShortestPath}($z_i^+$, $z_j^-$)\codecomment{set of the nodes} \label{line:path_finding}
        \STATE $\mathcal{P} \leftarrow \mathcal{P} \cup \mathcal{P}_{i,j}$
        \STATE $c_{i,j} \leftarrow | \mathcal{P}_{i,j} |$ \codecomment{construct a $K \times K$ cost matrix}
    \ENDFOR
\ENDFOR \label{line:cost_end}
\STATE $\mathcal{A} \leftarrow$ \textsc{Hungarian}($c_{i,j}$)\codecomment{$c_{i,j}$ is a cost matrix equivalent to a weighted bipartite graph for the Hungarian method} \label{line:hungarian}
\STATE $\mathcal{X}_\text{init} \leftarrow$ \textsc{RedistributionPlanning}($\mathcal{A}, \mathcal{Z}$) \label{line:redistribution_plan} 
\STATE $\mathcal{X} \leftarrow$ \textsc{Revising}($\mathcal{X}_\text{init}, \mathcal{P}, \mathcal{Z}$) \label{line:decompose_plan}
\STATE $\textsf{\small C1}, \textsf{\small C2}, \textsf{\small C3}, \textsf{\small C4} \leftarrow$ \textsc{CategorizeComponent}($\mathcal{X}, \mathcal{Z}$) \label{line:categorize}
\RETURN $\mathcal{X}$, $\textsf{\small C1}, \textsf{\small C2}, \textsf{\small C3}, \textsf{\small C4}$
}
\end{algorithmic}
\end{algorithm}

\begin{algorithm}[h]
\caption{{\scshape Revising}} \label{alg:Revising}
\begin{algorithmic}[1]
{\small
\floatname{algorithm}{Procedure}
\renewcommand{\algorithmicrequire}{\textbf{Input:} }
\renewcommand{\algorithmicensure}{\textbf{Output: }}
\REQUIRE $\mathcal{X}_\text{init}, \mathcal{P}, \mathcal{Z}$
\ENSURE Revised redistribution plan $\mathcal{X}$
\STATE $\mathcal{X} \leftarrow \emptyset$
\FOR{each flow $(z_i^+, z_j^-, N_{i,j})$ $\in \mathcal{X}_\text{init}$}
    \STATE $\mathcal{Z}_{i,j} \leftarrow$ \textsc{Decompose}($\mathcal{P}_{i,j}, \mathcal{Z}$) \codecomment{$\mathcal{Z}_{i,j}$ is a array include sequence of components (e.g., $(z_i^+, \cdots , z_j^-)$)} \label{line:breakdown}
    \FOR{$k=1$ to $|\mathcal{Z}_{i,j}|-1$} \label{line:start_decompose}
        \STATE Add $(\mathcal{Z}_{i,j}[k], \mathcal{Z}_{i,j}[k+1], N_{i,j})$ to $\mathcal{X}$
    \ENDFOR \label{line:end_decompose}
\ENDFOR
\STATE $\mathcal{X} \leftarrow$ \textsc{Merge}($\mathcal{X}$) \label{line:start_gathering}
\RETURN $\mathcal{X}$
}
\end{algorithmic}
\end{algorithm}

A travel distance can be obtained by finding a path on the roadmap. To construct a cost matrix, all paths between all pairs of surplus robots and tasks should be found, which incurs computational overheads. We reduce the computation by approximating robot-wise paths using component-wise paths. For each pair of an oversupplied component $z_i^+$ and an undersupplied component $z_j^-$, we find a shortest path $\mathcal{P}_{i,j}$, which consists of a set of nodes connecting a pair of component centers. A component center is defined as an intermediate node of a section or a JC node belonging to the component. We use the path length between $z_i^+$ and $z_j^-$ as the cost of any robot in $z_i^+$ performing any task in $z_j^-$. By this approximation, we can reduce the number of path planning queries in the roadmap. Lines~\ref{line:cost_start}--\ref{line:cost_end} of Alg.~\ref{alg:re-distribution} find the costs between oversupplied components and undersupplied components.

From an allocation $\mathcal{A}$ between the surplus robots in $z_i^+$ to the tasks in $z_j^-$ for all pairs of $i$ and $j$, we find the initial redistribution plan $\mathcal{X}_\text{init}$. In line~\ref{line:redistribution_plan} of Alg.~\ref{alg:re-distribution}, we search all assignments in $\mathcal{A}$ to find the number of robots moving from $z_i^+$ to $z_j^-$ denoted by $N_{ij}$. As a result, we obtain $\mathcal{X}_\text{init} =\{(z_i^+, z_j^-, N_{ij}) \mid i, j \text{ are respective component indices} \}$. Each tuple in $\mathcal{X}_\text{init}$ is called \textit{flow}, which describes the start and goal components with the quantity of robots to transfer between the components. 

The plan $\mathcal{X}_\text{init}$ can be executed by letting the robots follow the shortest paths $\mathcal{P}_{i,j}$ found in line~\ref{line:path_finding}. After executing the plan, the demand and supply can be balanced so $D_s=0$ for all components. However, $\mathcal{X}_\text{init}$ does not consider the traffic between components while the robots are redistributed. To mitigate this, we revise $\mathcal{X}_\text{init}$ to reduce the total travel distance of the robots, thereby lowering the overall traffic during redistribution.

\begin{figure}[h]
    \centering
    \captionsetup[subfigure]{justification=centering,singlelinecheck=false}
    \begin{subfigure}{0.235\textwidth}
        \includegraphics[width=\textwidth]{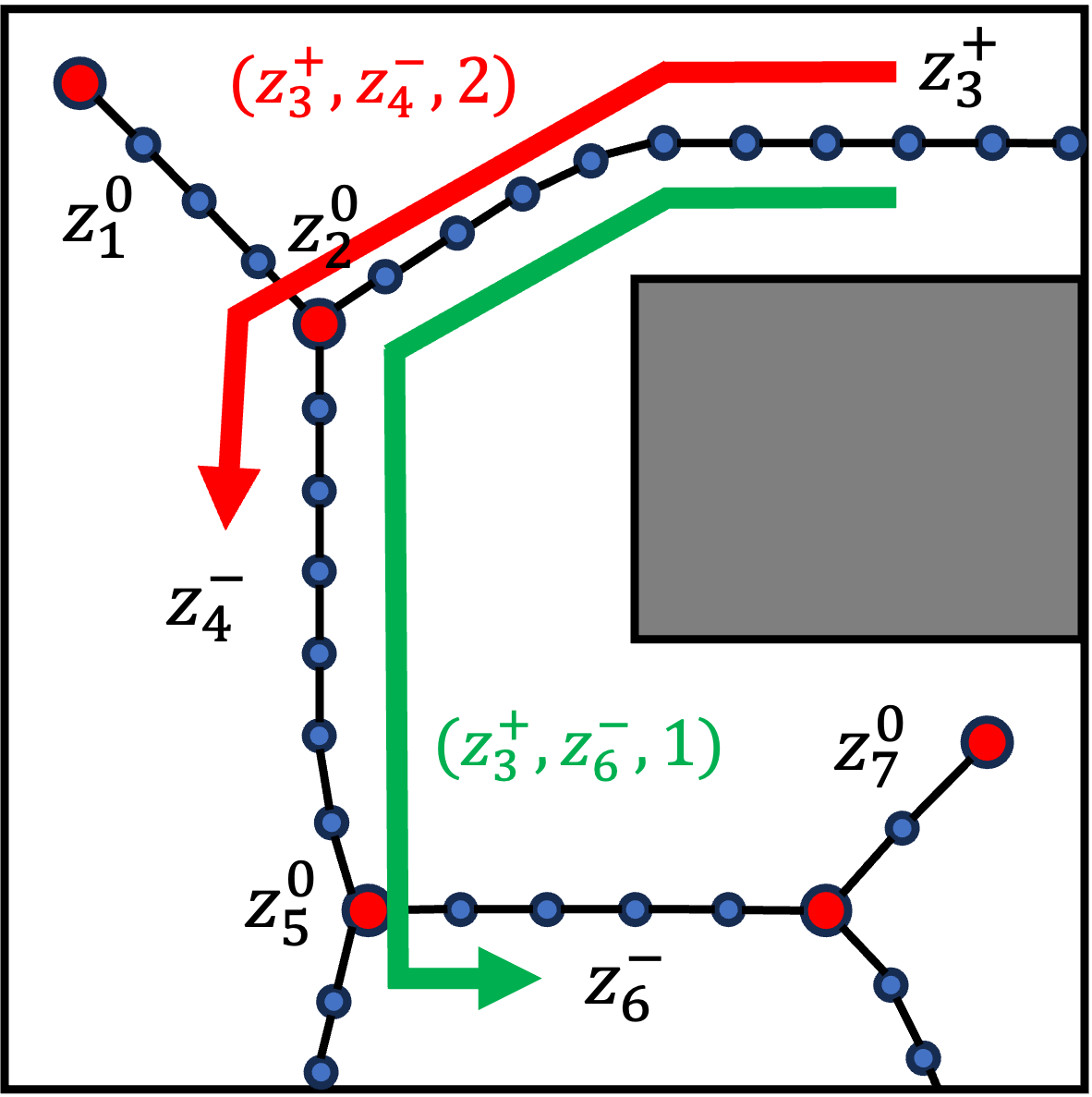}
        \caption{Initial redistribution plan $\mathcal{X}_\text{init}$}
        \label{fig:example1_1}
    \end{subfigure}
    \begin{subfigure}{0.235\textwidth}
        \includegraphics[width=\textwidth]{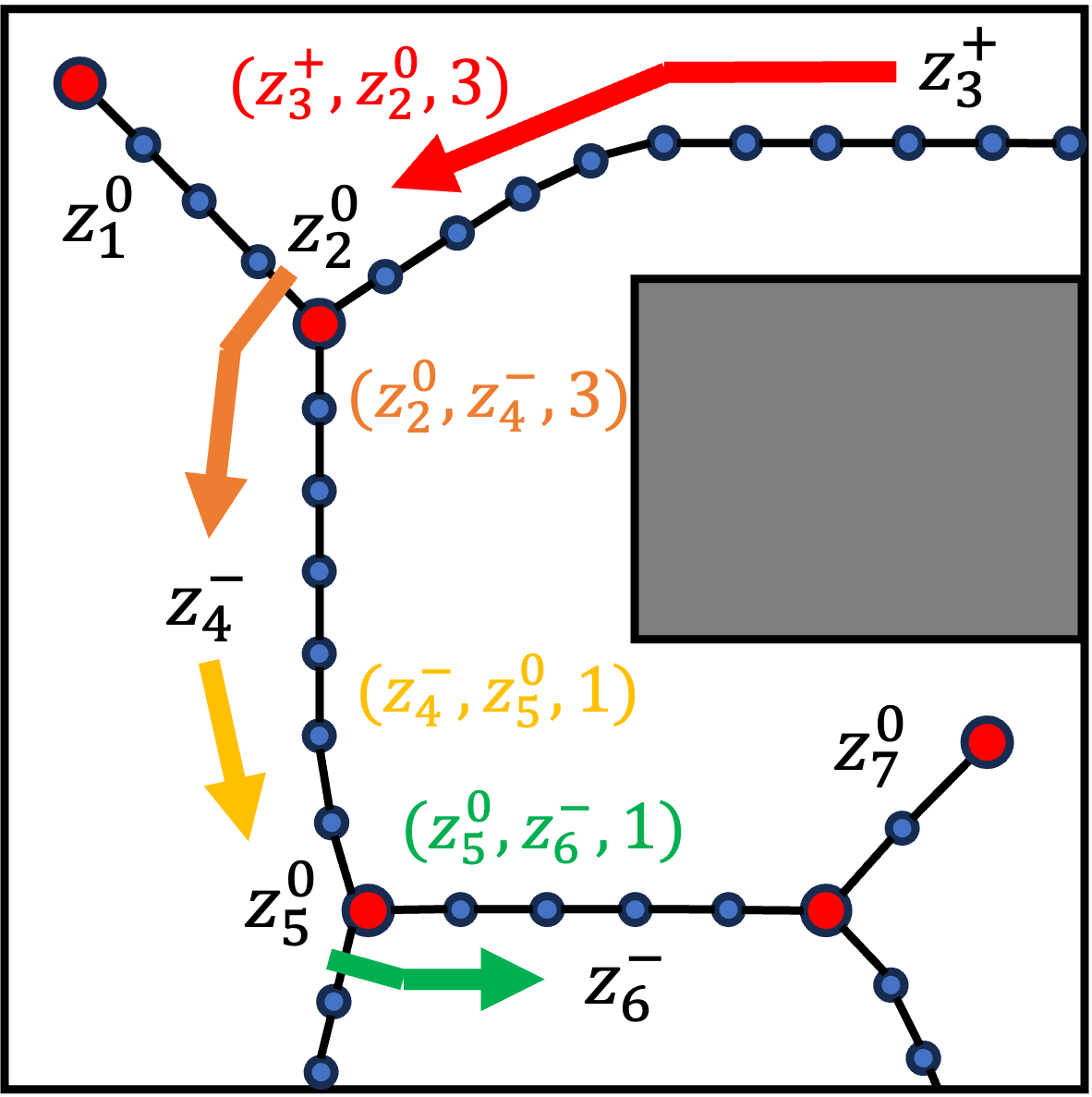}
        \caption{Revised redistribution plan $\mathcal{X}$}
        \label{fig:example1_2}
    \end{subfigure}
    \caption{Example of revising initial redistribution plan to the final redistribution plan. After the revision, flows are broken down into component-wise flows.}
    \label{fig:example1}
    \vspace{-10pt}
\end{figure}

For example, an initial plan $\mathcal{X}_\text{init} = \{(z_3^+, z_4^-, 2)$, $(z_3^+, z_6^-, 1)\}$ in Fig.~\ref{fig:example1_1} shows that total three robots move out from $z_3^+$ to other components. One of the three robots reaches $z_6^-$ while two others stop at $z_4^-$. If there exists at least one robot in $z_4^-$, one of them can be sent to $z_6^-$ so all three robots from $z_3^+$ can stop at $z_4^-$. All of the flow can be revised into adjacent component-wise flow. This revision reduces the maximum number of traversed components of the robots from four to one. But if there is no robot associated with $z_2^\circ$ and $z_4^-$, one of the three robots from $z_3^+$ will be directly sent to $z_5^\circ$. The revised plan $\mathcal{X} = \{(z_3^+, z_2^\circ, 3)$, $(z_2^\circ, z_4^-, 3)$, $(z_4^-, z_5^\circ, 1)$, $(z_5^\circ, z_6^-, 1)\}$ is described in Fig.~\ref{fig:example1_2}.

The revision process presented in Alg.~\ref{alg:Revising} uses $\mathcal{P}_{i,j}$, $\mathcal{Z}$, and $\mathcal{X}_\text{init}$. The robots moving according to this plan are assumed to follow $\mathcal{P}_{i,j}$. From $\mathcal{P}_{i,j}$, we can determine a sequence of components, denoted by $\mathcal{Z}_{i,j}$, that $\mathcal{P}_{i,j}$ passes (line~\ref{line:breakdown}). In other words, $\mathcal{Z}_{i,j}$ represents the components to pass by robots if a flow in $\mathcal{X}_\text{init}$ is executed. In the example, $(z_3^+, z_6^-, 1)$ in Fig.~\ref{fig:example1_1} has $\mathcal{Z}_{3,6} = (z_3^+, z_2^\circ, z_4^-, z_5^\circ, z_6^-)$, which indicates that the path from $z_3^+$ to $z_6^-$ comprise four path segments where one from $z_3^+$ to $z_2^\circ$, from $z_2^\circ$ to $z_4^-$, from $z_4^-$ to $z_5^\circ$, and another from $z_5^\circ$ to $z_6^-$. Based on $\mathcal{Z}_{3,6}$, $(z_3^+, z_6^-, 1)$ is split into $(z_3^+, z_2^\circ, 1)$, $(z_2^\circ, z_4^-, 1)$, $(z_4^-, z_5^\circ, 1)$, and $(z_5^\circ, z_6^-, 1)$ (lines~\ref{line:start_decompose}--\ref{line:end_decompose}). This split is repeated for all flows in $\mathcal{X}_\text{init}$. All resulting flows are inserted into $\mathcal{X}$. 

If there are multiple flows with the same start and end components, they are merged by adding up the number of transferred robots with the same start and end. In the running example, $\mathcal{X}= ((z_3^+, z_2^\circ, 2)$, $(z_2^\circ, z_4^-, 2)$, $(z_3^+, z_2^\circ, 1)$, $(z_2^\circ, z_4^-, 1)$, $(z_4^-, z_5^\circ, 1)$, $(z_5^\circ, z_6^-, 1))$ has two flows sending robots from $z_3^+$ to $z_2^\circ$ and from $z_2^\circ$ to $z_4^-$. After the merger, we have $((z_3^+, z_2^\circ, 3)$, $(z_2^\circ, z_4^-,3)$, $(z_4^-, z_5^\circ, 1)$, $(z_5^\circ, z_6^-, 1))$ (line~\ref{line:start_gathering}).

\subsection{Prioritized flow execution for robot redistribution}
\label{sec:flow}

The flows in $\mathcal{X}$ do not specify which flow is executed first. Also, they do not determine which robots perform which tasks but only the amount to be transferred. Thus, we further process $\mathcal{X}$ to execute the redistribution plan and find an allocation of robots to tasks. For this process, we introduce four categories of components \textsf{\small C1} to \textsf{\small C4} (line~\ref{line:categorize} of Alg.~\ref{alg:re-distribution}) according to their characteristics with respect to the patterns of robot flows in the components, as summarized in Table~\ref{tab:category}.

Since components that belong to \textsf{\small C1} do not have any incoming and outgoing robots, $\mathcal{X}$ has no \textsf{\small C1} component. \textsf{\small C2} components have only outgoing robots, so they appear only as the start component in $\mathcal{X}$. Similarly, \textsf{\small C3} components have only incoming robots and appear only as the end component in $\mathcal{X}$.  Lastly, \textsf{\small C4} components have both incoming and outgoing robots, so they appear in $\mathcal{X}$ as both the start and end components. Therefore, \textsf{\small C2} and \textsf{\small C3} components work as sources and sinks, respectively. \textsf{\small C4} components manage the traffic from the sources to sinks. In the running example (Fig.~\ref{fig:example1}) where we have $\mathcal{X} = ((z_3^+, z_2^\circ, 3)$, $(z_2^\circ, z_4^-,3)$, $(z_4^-, z_5^\circ, 1)$, $(z_5^\circ, z_6^-, 1))$, $z_3^+$ is a \textsf{\small C2} component since it works as the start component only (so in the first place in the tuples). On the other hand, $z_6^-$ is a \textsf{\small C3} component while $z_2^\circ$, $z_4^-$, and $z_5^\circ$ is a \textsf{\small C4} component. The rest of the components do not appear (i.e., $z_1^\circ, z_7^\circ$) belong to \textsf{\small C1}.

\begin{table}[h]
\centering
\caption{Characteristics of categories that distinguish components.}
\begin{tabular}{|l|l|l|l|l|}
\hline
Category & \textsf{C1} & \textsf{C2} & \textsf{C3} & \textsf{C4} \\ \hline
Incoming robot? & $\times$ & $\times$ & \checkmark & \checkmark \\ \hline
Outgoing robot? & $\times$ & \checkmark & $\times$ & \checkmark \\ \hline
\end{tabular}
\label{tab:category}
\end{table}

From $\mathcal{X}$ and the categorization, we perform a planning to find a sequence of nodes for each robot, where the nodes work as waypoints to guide the robot to move along. Then, the robots that reach their designated components are allocated to tasks in the respective components.

We first determine the priority of the flows in $\mathcal{X}$. Then, we move the robots hypothetically as a planning process to find the node sequences of all robots. We determine the priority of flows based on the categories of the start components in the flows. Since \textsf{\small C1} and \textsf{\small C3} components are never the start components owing to their characteristics (i.e., both do not send any robot), we only consider the flows starting from \textsf{\small C2} and \textsf{\small C4} components. Since \textsf{\small C4} components have both incoming and outgoing robots, the flows starting from \textsf{\small C4} components are executed later once they receive all robots from other components. 

Sending robots to \textsf{\small C4} should precede to reduce the makespan of the system since the robots sent to \textsf{\small C4} have the potential to be further transferred to other components but \textsf{\small C3} are not. Therefore, we execute\footnote{As noted, the execution is done hypothetically as a planning process.} the flows moving from \textsf{\small C2} to \textsf{\small C4} components first. We search $\mathcal{X}$ to find such flows and execute them sequentially in turn as soon as each of them is found.

To prevent deadlocks when the robots actually move, it is also important to determine which robot should be sent and also the arrival order of robots in each receiving component. When deciding which robot to send from a starting component, we consider the configuration of the roadmap. If the starting node of this component is directly connected to the destination component, the robot closest to the starting node is selected to move first. If the connection is from the end node of the starting component to the destination component, then the robot closest to the end node will move first. Additionally, among all robots arriving at a receiving component, the closest robot to the component should arrive first. Once all flows from \textsf{\small C2} to \textsf{\small C4} components are executed, the flows from \textsf{\small C2} to \textsf{\small C3} components are searched and executed.

Subsequently, we search $\mathcal{X}$ to execute the flows from a \textsf{\small C4} component to another \textsf{\small C4}. While sequentially executing the flows, some \textsf{\small C4} components could still be receiving robots from \textsf{\small C4} components. They are not chosen to be executed until they completely receive all robots. After all flows between \textsf{\small C4} components are accomplished, the flows from \textsf{\small C4} components to \textsf{\small C3} components are executed. When a \textsf{\small C4} component sends robots to other components, the robots that belong to the component from the beginning are chosen in the same way as we choose which robot to be sent in the \textsf{\small C2} components. If all such robots are depleted but more robots still have to be sent, the robots from other components are chosen to be sent in the order of the distance that they traveled on the roadmap.

\subsection{Task assignment within components}
\label{sec:assign}

After executing all of the flows in the redistribution plan $\mathcal{X}$ hypothetically, we are aware of the destination component for every robot. Consequently, the assignment between the robots and tasks existing in each component is determined based on the arrival order of the robots, with the robot with the shorter traveling distance being allocated to the task further from the entering JC node\footnote{As we assumed, robot kinodynamics constraints are not considered so the arrival time can be computed from the travel distance and velocity.}. This first-come first-serve allocation prevents the robot from blocking the pathway of the robots arriving later to their tasks. If more than one robot arrives at the node at the same time, the tie can be broken randomly. 

\begin{figure}[h]
    \centering
    \begin{subfigure}{0.4\textwidth}
        \includegraphics[width=\textwidth]{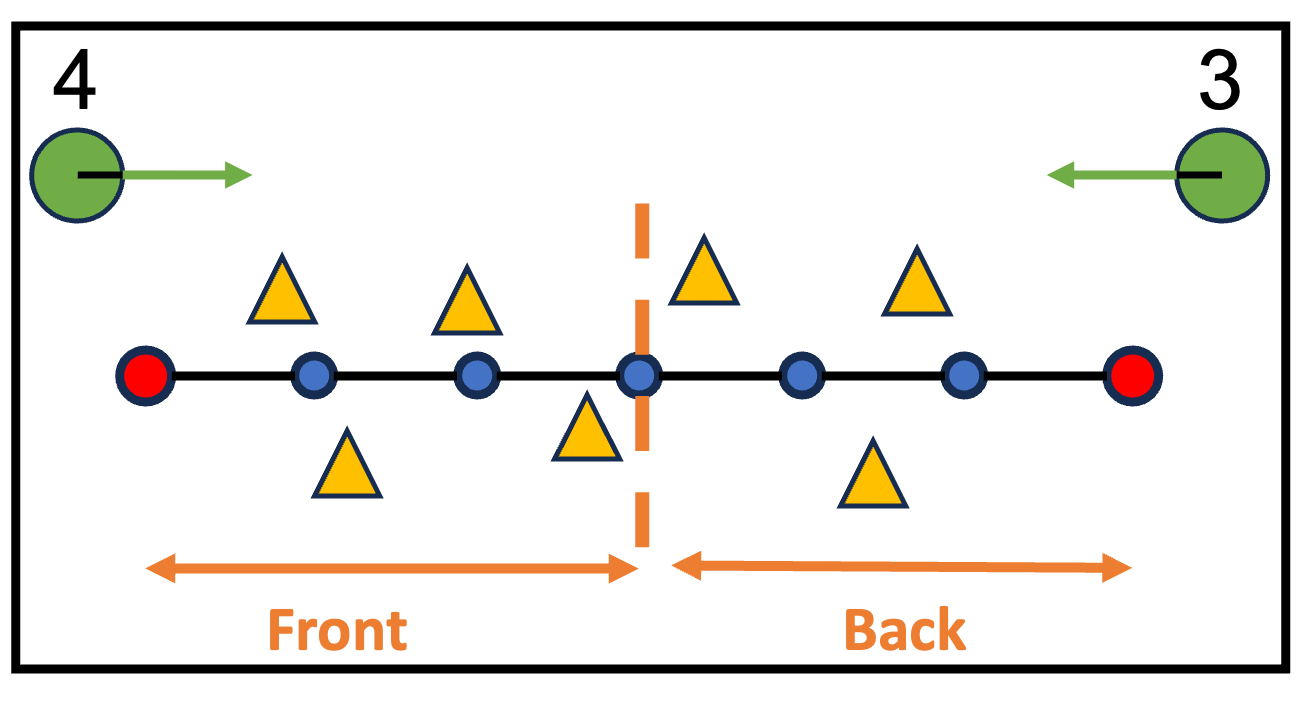}
        \caption{Task grouping when robots enter from both ends. Tasks are split into two groups (front and back) based on the direction of robot entry.}
        \label{fig:task_group_1}
    \end{subfigure}
    \hfill
    \begin{subfigure}{0.4\textwidth}
        \includegraphics[width=\textwidth]{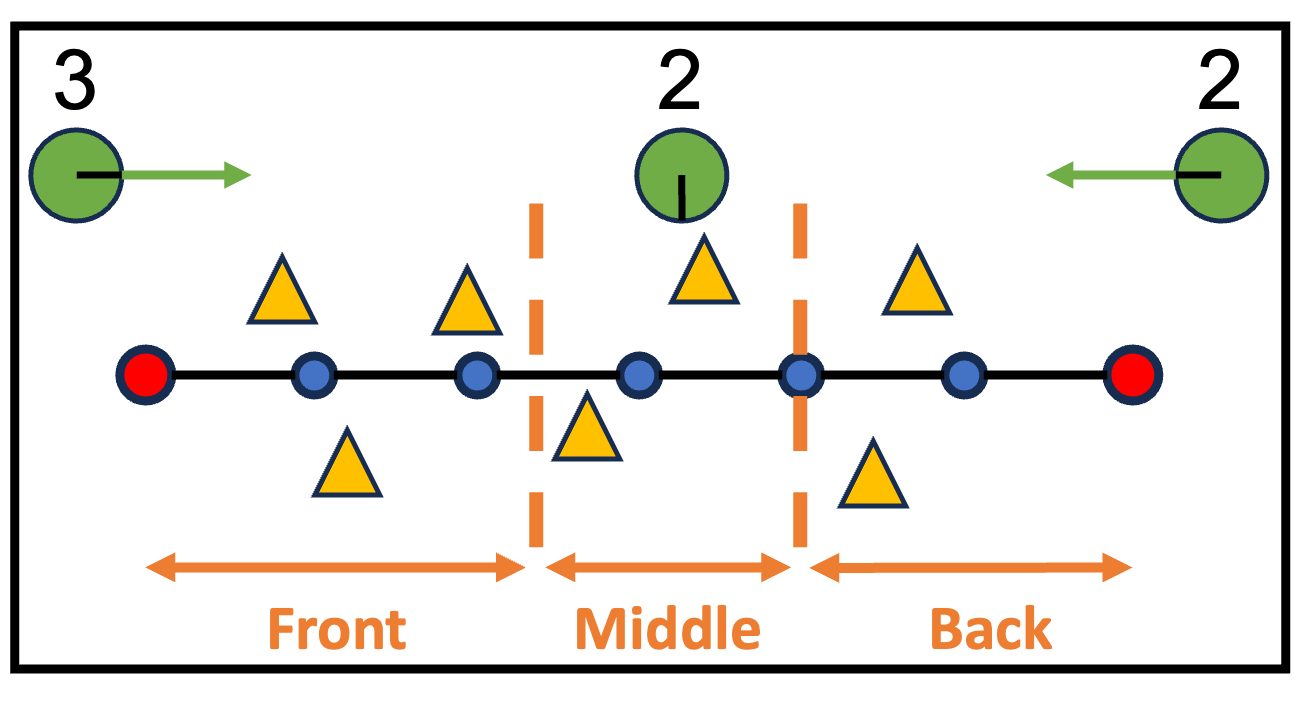}
        \caption{Task grouping when both incoming and native robots are present. Tasks are split into three groups (front, middle, and back) to reduce cross-traffic and enable smooth task assignment.}
        \label{fig:task_group_2}
    \end{subfigure}
    \vspace{-5pt}
    \caption{Example of task grouping in a section. The yellow triangles denote tasks associated with the section, and green circles denote robots. Also, green arrows indicate incoming robot directions. Orange dashed lines show task group boundaries.}
    \label{fig:task_group}
    \vspace{-10pt}
\end{figure}

There could be robots entering into a section from the both sides. If the robots are assigned to the tasks that are closer to their entering side, the traffic from both sides could not be mingled, which incurs less conflicts. Particularly, we divide the tasks in the section into two groups based on the index they have in the section (i.e., front, and back groups as shown in the orange color in Fig.~\ref{fig:task_group_1}) according to the number of arriving robots from each side. Therefore, the robots coming from the start node of the section are allocated to the front task group and vice versa. If there are some robots that reside in the section without being transferred, the tasks are rather divided into three groups (i.e., front, middle, and back task groups illustrated in Fig.~\ref{fig:task_group_2} as the orange color). Thus, the robots entering from the start and end nodes of the section are assigned to the front and back task groups, respectively. The robots residing in the section are assigned to the middle task group, starting with the robot and task each having the lowest index to the highest index.

When the section receives the robot for only one side and some robots reside in the section, we also divide the tasks into two groups. On the other hand, when the section only has a robot received in one direction or only a robot that resides in the section we divide the tasks into only one group. Additionally, the tasks associated with the JC node are served as one group.

Finally, in our method, we provide each robot with a recommended path consisting of only JC nodes derived from the node sequence on the roadmap. This path is used as a reference to guide the robot to its assigned task while enabling it to autonomously navigate in continuous space using its own local controller. Therefore, the robots do not strictly follow all nodes on the roadmap but use the JC-node-based reference waypoints to avoid obstacles and congestion in a feasible trajectory.

\subsection{Analysis}
\label{sec:prove}

\thm \textbf{4.1.} For a JC node that has at least one robot to pass, the robot(s) passes that JC node in one direction.

\pf.
As a base case, we consider two sections $z$ and $z^\prime$ that are adjacent, so one JC node connects them. Let $\mathcal{X}_\text{init}$ be the redistribution plan obtained from an allocation $\mathcal{A}$ obtained in Sec.~\ref{sec:rm}. From the revision process of $\mathcal{X}_\text{init}$ in Sec.~\ref{sec:rm}, we have $\mathcal{X}$ with flows distributing robots between adjacent components only. 

Suppose that the redistribution plan instructs the robots to cross the JC node connecting $z$ and $z^\prime$ from the opposite direction. In other words, $\mathcal{X}$ has flows that move robots from $z$ to $z^\prime$ through the JC node and other flows that move robots from $z^\prime$ to $z$ through the same JC node. These flows lead to the result of sending robots from $z$ to $z^\prime$ and also from $z^\prime$ to $z$. After this redistribution, tasks in both components can be fulfilled by the robots came from each other components. Let the cost of this allocation be $c$.

On the other hand, we can have another allocation with a cost cheaper than $c$ by performing tasks by the robots in the same component. Since $\mathcal{X}$ is a consequence of $\mathcal{A}$ minimizing the cost of the task allocation, $\mathcal{X}$ cannot have such flows in the opposite direction producing a higher cost. Thus, the supposition, that the robots cross a JC node from the opposite direction, is contradicted.

\begin{figure}[h]
    \centering
    \includegraphics[width=0.15\textwidth]{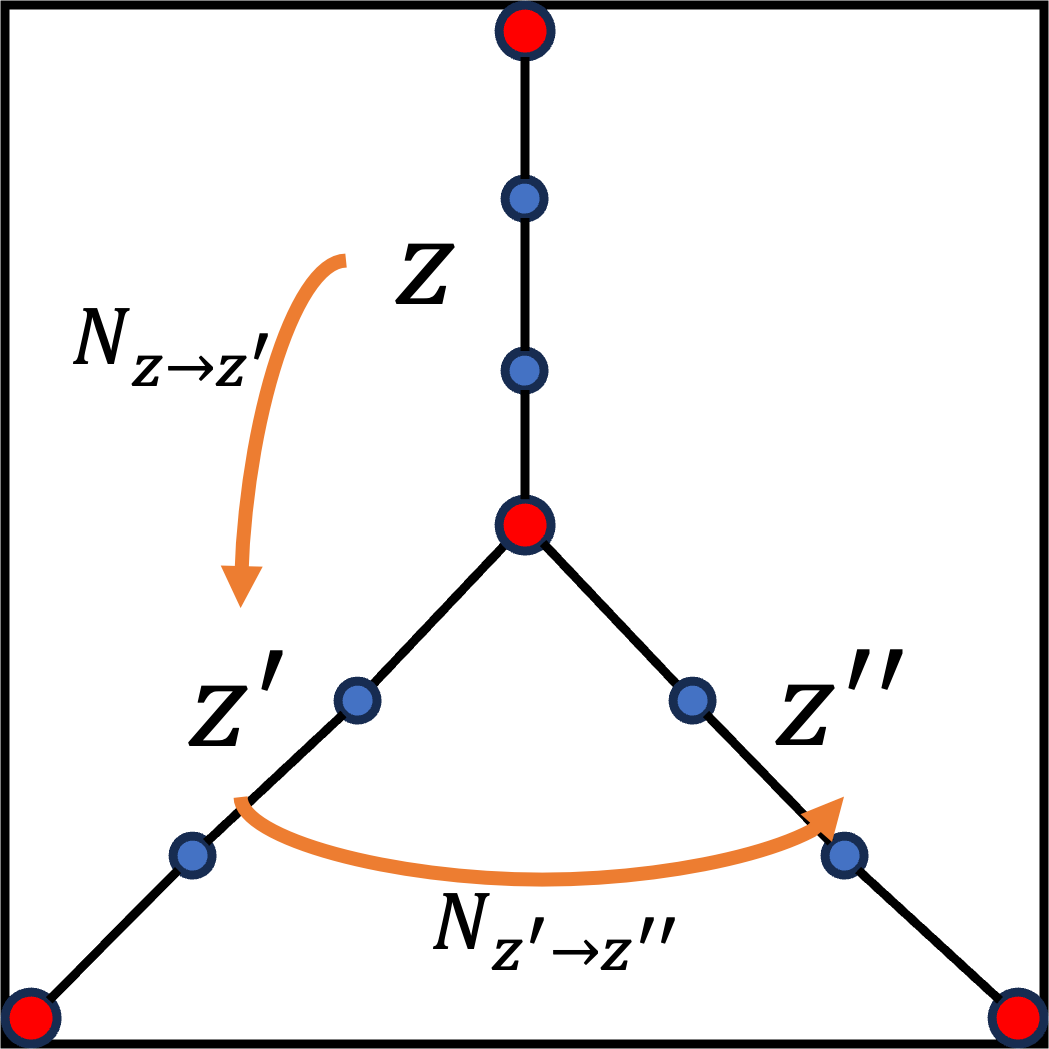}
    \caption{Example environment for Theorem 4.1}
    \label{fig:lemma4_1}
    \vspace{-10pt}
\end{figure}

Let us consider another case where three adjacent sections $z$, $z^\prime$, and $z^{\prime\prime}$ are connected to one JC node as shown in Fig.~\ref{fig:lemma4_1}. Suppose that $\mathcal{X}$, computed from $\mathcal{A}$, has flows that move $N_{z \rightarrow z^\prime} \in \mathbb{N}$ robots from $z$ to $z^\prime$ through the JC node and other flows that move $N_{z^\prime \rightarrow z^{\prime\prime}} \in \mathbb{N}$ robots from $z^\prime$ to $z^{\prime\prime}$ through the same JC node. In this case, we can have another allocation in which the robots travel shorter. Specifically, when $N_{z \rightarrow z^\prime}=N_{z^\prime \rightarrow z^{\prime\prime}}$, directly transferring $N$ robots from $z$ to $z^{\prime\prime}$ can reduce the cost. For $N_{z \rightarrow z^\prime} > N_{z^\prime \rightarrow z^{\prime\prime}}$, the cost can be cheaper when transferring $N_{z \rightarrow z^{\prime}}-N_{z^\prime \rightarrow z^{\prime\prime}}$ robots from $z$ to $z^\prime$ and $N_{z^\prime \rightarrow z^{\prime\prime}}$ robots from $z$ to $z^{\prime\prime}$. Conversely, if $N_{z \rightarrow z^\prime} < N_{z^\prime \rightarrow z^{\prime\prime}}$, it is more cost-effective to move $N_{z \rightarrow z^\prime}$ robots from $z$ to $z^{\prime\prime}$ and $N_{z^\prime \rightarrow z^{\prime\prime}} - N_{z \rightarrow z^\prime}$ robots from $z^\prime$ to $z^{\prime\prime}$. Since $\mathcal{A}$ minimizes the cost for allocation, the supposition contradicts as there could be allocations with lower costs.

In summary, if a JC node passes at least one robot to a section, the section received robots must not send robots to other sections through that JC node. Thus, there is no pair of sections where the robots cross a JC node in the opposite direction. 
\qed
\smallskip

\thm \textbf{4.2.} For every edge in the roadmap, no robot crosses the edge in the opposite direction.

\pf. As described in Sec.~\ref{sec:assign}, we group tasks in a section into one, two (i.e., front, back) or three (i.e., front, middle, back). Then robots from other components and \textit{native} robots, which are not from any other component, are allocated to the respective task group by the case.

When the robots come from only one side, the movement of the robots is naturally unidirectional. Furthermore, when the robots enter from both sides of the section, the tasks are bisected, so the moves of robots toward the front and back task groups are also reduced to the first case with unidirectional moves only.

For native robots, we follow an index-based ordering strategy described in Sec.~\ref{sec:demand}, where the remaining robots are allocated to tasks sequentially from the lowest-indexed robot to the lowest-indexed task. This index reflects the spatial order of nodes in the section. Although robots may move in different directions within the section depending on their respective task positions, the allocation ensures that their paths do not intersect or block one another. This is because robots and tasks are assigned in order based on their spatial indices, so that each robot is matched to the closest unassigned task in sequence. As a result, even if the travel directions differ, the resulting paths remain disjoint, preserving the non-interference property across all edges in the section, including those connected to JC nodes.

\qed
\smallskip

\thm \textbf{4.3.} When robots travel to their assigned tasks along the reference waypoints on the roadmap, those arriving first do not block the way for others.

\pf. 
We assume that the arrival order of the robot, which is calculated by the traveled distance, is followed. Under the assumption, any pair of robots from the same side does not block each other owing to the first-come first-serve allocation making the robot that arrives first move to the farthest task in the respective task group (i.e., front or back task group).

Furthermore, for any native robot, while other native robots or tasks may be located between the robot and its assigned task in the section, there is no assigned robot-task pair whose elements both lie between that robot and its destination. This is because native robots and tasks are assigned in index order, ensuring that no later-assigned pair can obstruct the path of an earlier-assigned one. Thus, native robots do not block each other.

In addition, for each JC node, the robots that arrive first are transferred to the adjacent sections. The remaining robots, which arrive later in a quantity matching the number of tasks associated with the JC node, are then allocated to these tasks. Therefore, the remaining robots in each JC node never block other robots from moving.
\qed
\smallskip

\thm \textbf{4.4.} Flow execution in MRTA-RM is complete under a connected GVD-based roadmap.

\pf.
We prove the completeness of the flow execution in Sec.~\ref{sec:flow}. To begin, we consider that flows starting from components in \textsf{\small C2} are executed first. Since these flows do not require any conditions, they can always proceed without limitations.

However, for flows starting from components in \textsf{\small C4}, all robots must have been received before the flow can be executed. Therefore, to prove completeness, we demonstrate that the system cannot reach a state in which none of the remaining flows can satisfy their execution conditions.

Assume that during the flow execution, the system reaches a state in which none of the remaining flows can satisfy their conditions for execution. Let the set of flows that remain unexecuted at this stage be denoted by $\mathcal{F}_l$, and let the set of corresponding starting components be denoted by the $\mathcal{Z}_l$. In this situation, every component in $\mathcal{Z}_l$ has not fully received robots.

In this scenario, since all of the flows from \textsf{\small C2} components are executed, there is at least one flow that is not included in $\mathcal{F}_l$ and is heading to a component in $\mathcal{Z}_l$. This implies that at least one component in \textsf{\small C4} is not included in $\mathcal{Z}_l$ and has not met the execution condition. This contradicts the initial state. Thus, we conclude that no such state can exist, and all flows will eventually be executed, ensuring the completeness of the flow execution.
\qed
\smallskip

By Theorems 4.1--3, Properties 1 and 2 in Sec.~\ref{sec:description} are satisfied if the robots can exactly follow the paths along the roadmap. In other words, the robots do not move in the opposite direction and no robot blocks other moving robots. Thus, the robots can travel to their allocated tasks without conflicts. However, conflict-free paths could not be guaranteed if the robots deviate from the paths from the roadmap. Such deviations may occur if the robots choose alternative routes not aligned with the planned flows, which can invalidate the assumptions used during flow execution and task allocation, such as travel distances and entering directions. To mitigate this, either the robots should be guided to follow the computed reference waypoints, or conflict-aware motion planners such as MAPF solvers can be employed to compute paths that preserve the deadlock-free properties.
\smallskip

\thm \textbf{4.5.} MRTA-RM executes in polynomial time.

\pf
We analyze the computational complexity of the MRTA-RM in terms of the number of robots $N$, the number of components $\vert \mathcal{Z} \vert$, and the number of nodes $\vert \mathcal{V} \vert$ in the roadmap $G=(\mathcal{V},\mathcal{E})$.

The demand and supply analysis in Sec.\ref{sec:demand} (line~\ref{line:dns_analysis} in Alg.~\ref{alg:re-distribution}) has a time complexity of $O(N\vert \mathcal{V}\vert)$ as it involves finding the closest node in $G$ for each robot and task.

In the robot redistribution plan (Sec.\ref{sec:rm}), to construct the cost matrix in lines~\ref{line:cost_start}-\ref{line:cost_end} of Alg.~\ref{alg:re-distribution}, the Dijkstra algorithm is employed to find the shortest paths between oversupplied components and undersupplied components, resulting in time complexity of $O(\vert \mathcal{V} \vert^2)$. In this part, $\vert \mathcal{Z}\vert$ is smaller than $\vert \mathcal{V}\vert$ since the sections contain several vertices. Accordingly, we compress $G$ into a component-wise graph, reducing the time complexity to $O(\vert \mathcal{Z} \vert^2)$.

For an allocation $\mathcal{A}$, we employ the Hungarian method, which computes in $O(N^3)$. To make an initial robot redistribution plan $\mathcal{X}_\text{init}$ from $\mathcal{A}$, we determine which robots in oversupplied components are assigned to undersupplied components for each allocated pair in $\mathcal{A}$. Therefore, line~\ref{line:redistribution_plan} in Alg.~\ref{alg:re-distribution} has a complexity of $O(N)$. The revision process in Alg.\ref{alg:Revising}, which transforms $\mathcal{X}_\text{init}$ into $\mathcal{X}$, involves decomposing and merging sequences. Decomposing each flow in $\mathcal{X}_\text{init}$ has a complexity of $O(N\vert \mathcal{Z}\vert^2)$, where $\vert \mathcal{Z}\vert^2/2$ represents the number of potential connections between components. Similarly, merging sequences also requires $O(N\vert \mathcal{Z}\vert^2)$ time for up to $O(N\vert \mathcal{Z}\vert^2)$ fragments of decomposed redistribution plans. In line~\ref{line:categorize}, we search for each component to determine if there are incoming or outgoing flows from it in $\mathcal{X}$. Consequently, this process has a time complexity of $O(\vert \mathcal{Z}\vert^3)$.

Lastly, in the flow execution step (Sec.\ref{sec:flow}), flows from \textsf{\small C2} are executed unconditionally, whereas flows from \textsf{\small C4} require a condition to be met. Nevertheless, since there is always at least one feasible flow (Theorem 4.4), this can be done in $O(\vert \mathcal{Z}\vert^3)$ time, as it is completed in $\vert \mathcal{Z}\vert$ iterations for at most $\vert \mathcal{Z}\vert^2/2$ flows.

In summary, the MRTA-RM has a computational complexity of $O(N\vert \mathcal{V}\vert+\max(N^3, \vert \mathcal{Z}\vert^3))$.
\qed
\smallskip

\section{Experiments}
\begin{figure}[]
    \centering
    \includegraphics[width=0.48\textwidth]{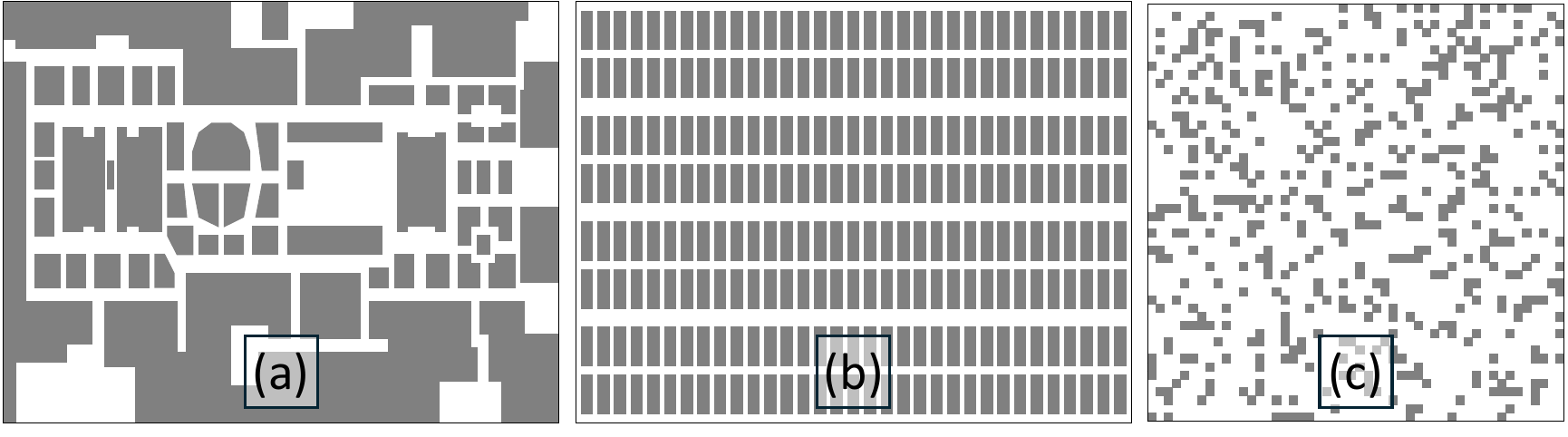}
    \caption{Environments used for evaluation: (a) Shopping mall ($1760\times900$ units), (b) Warehouse ($2200\times880$ units), (c) Cluttered (1000$\times$1000 units)}
    \label{fig:full_envs}
    \vspace{-5pt}
\end{figure}

We have three different environments that have different configurations as shown in Fig.~\ref{fig:full_envs}. The first environment \textit{shopping mall} is from the floorplan of a department store with halls and corridors. It has many long corridors without exits which could incur congestion and deadlocks. The \textit{warehouse} environment is with many intersections where robots could frequently encounter. The \textit{clutter} environment is with randomly distributed square obstacles with the same size. The sizes of the environments are $1760\times900$, $2200\times880$, and $1000\times1000$ units, respectively. The shape of the robot is circular whose radius is six units. In those environments, we have two scenarios depending on the distributions of the robots and tasks. The \textit{random} scenario have robots and tasks distributed randomly in the free space. The \textit{separated} scenario has robots in the left side of the environment while tasks are in the right side. The latter scenario incurs more conflicts between robots as the robots form large chunks.

\begin{figure}[t!]   
    \centering
    \includegraphics[width=0.48\textwidth]{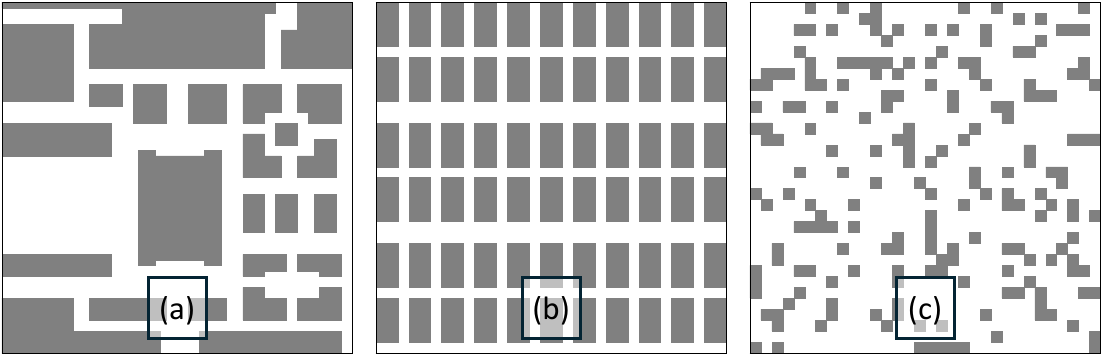}
    \caption{Part of the environments for small-scale experiments comparing CBS-based methods. The size of all environments is $640\times640$ units.}
    \label{fig:mini_envs}
    \vspace{-15pt}
\end{figure}

We have two different sets of tests. The first set (Sec.~\ref{sec:small}) compares our proposed method with other ones using CBS to resolve conflicts between robots. The search-based methods explicitly resolve the conflicts which require additional computations so they often do not scale with the number of robots. Thus, we run those methods with 30 robots at most. To keep the density of the robots high, we use only small part of the three environments as shown in Fig.~\ref{fig:mini_envs}. The second set (Sec.~\ref{sec:large}) compares our method with the methods that do not consider conflicts in task allocation. Since those methods are simple and fast, we scale up the problem instances up to 500 robots using the complete maps of the three environments shown in Fig.~\ref{fig:full_envs}, which we refer to as the full environments. These are in contrast to the partial or cropped regions used for small-scale tests shown in Fig.~\ref{fig:mini_envs}.

In order to take the inter-robot conflicts into account in the tests, we use a dynamic robotic simulator. We use Unity Engine where the navigation of robots is done by Unity NavMesh which enables the robots to find paths and navigate along the paths. The robots allocated to the tasks navigate in the environments where their dynamics (e.g., acceleration, deceleration) are properly modeled.  Each robot runs a built-in local collision avoidance that uses the Reciprocal Velocity Obstacles (RVO) algorithm~\cite{RVO}. Although RVO can handle local collisions, it cannot resolve a deadlock situation in narrow passages. 

The performance metrics of the tests are the computation success rate, computation time, success rate, makespan, and the sum of costs (SoC). The computation success rate counts the number of instances solved within the 5-minute time limit. The computation time represents the average elapsed time to solve the successful instances. Thus, there could be a bias in the computation time if successful instances are scarce. The success rate counts the number of successful executions of the task allocation without deadlocks. In other words, the computation success rate measures the computational efficiency of the method while the success rate measures the quality of allocation  in terms of preventing deadlocks. A deadlock is defined as the situation where any of the robots gets stuck for five seconds. The makespan measures the total elapsed time to achieve all tasks by the robots. This is our main objective value to be optimized. We also measure the SoC which is the sum of travel times of all robots.

All experiments are done with a system with an AMD 5800X 3.8GHz CPU, 32GB RAM, and Python 3.11.



\begin{figure*}[!h]
    \centering
    \includegraphics[width=0.9\textwidth]{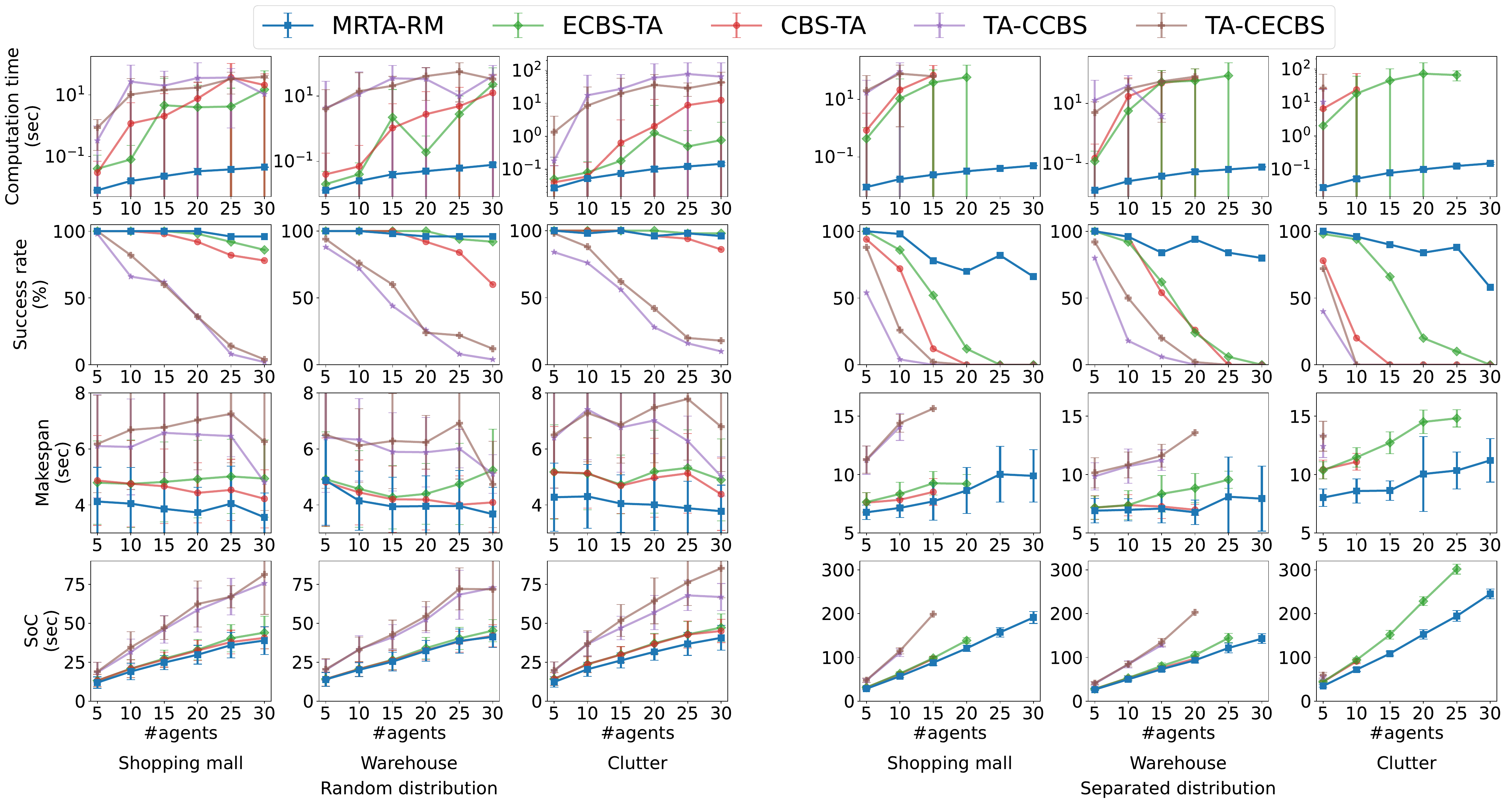}
    \vspace{-10pt}
    \caption{The comparison result with the methods that explicitly resolve conflicts. Success rate here refers to the percentage of instances in which the method successfully computed a solution within the computation time limit and completed the assigned tasks in dynamic simulation without encountering any deadlock. Our proposed method outperforms all the compared methods in all metrics.}
    \label{fig:mini-exp}
    \vspace{-5pt}
\end{figure*}

\begin{table*}[!h]
\centering
\caption{The comparison result with the methods that explicitly resolve conflicts. Standard deviations are omitted owing to the space limit.}
\vspace{-5pt}
\label{tab:small_comp}
\resizebox{\textwidth}{!}{%
\begin{tabular}{|cc||ccccccccccccccc|}
\hline
\multicolumn{2}{|c|}{Distribution} &
  \multicolumn{15}{c|}{Random} \\ \hline
\multicolumn{2}{|c|}{Method} &
  \multicolumn{3}{c|}{CBS-TA} &
  \multicolumn{3}{c|}{ECBS-TA} &
  \multicolumn{3}{c|}{TA-CCBS} &
  \multicolumn{3}{c|}{TA-CECBS} &
  \multicolumn{3}{c|}{MRTA-RM} \\ \hline
\multicolumn{2}{|c|}{\#Robots} &
  \multicolumn{1}{c|}{10} &
  \multicolumn{1}{c|}{20} &
  \multicolumn{1}{c|}{30} &
  \multicolumn{1}{c|}{10} &
  \multicolumn{1}{c|}{20} &
  \multicolumn{1}{c|}{30} &
  \multicolumn{1}{c|}{10} &
  \multicolumn{1}{c|}{20} &
  \multicolumn{1}{c|}{30} &
  \multicolumn{1}{c|}{10} &
  \multicolumn{1}{c|}{20} &
  \multicolumn{1}{c|}{30} &
  \multicolumn{1}{c|}{10} &
  \multicolumn{1}{c|}{20} &
  30 \\ \hline
\multicolumn{1}{|c|}{\multirow{3}{*}{\begin{tabular}[c]{@{}c@{}}Comp.\\ Success rate\\ (\%)\end{tabular}}} &
  Shopping mall &
  \multicolumn{1}{c|}{\textbf{100}} &
  \multicolumn{1}{c|}{92} &
  \multicolumn{1}{c|}{78} &
  \multicolumn{1}{c|}{\textbf{100}} &
  \multicolumn{1}{c|}{98} &
  \multicolumn{1}{c|}{86} &
  \multicolumn{1}{c|}{66} &
  \multicolumn{1}{c|}{36} &
  \multicolumn{1}{c|}{2} &
  \multicolumn{1}{c|}{82} &
  \multicolumn{1}{c|}{36} &
  \multicolumn{1}{c|}{4} &
  \multicolumn{1}{c|}{\textbf{100}} &
  \multicolumn{1}{c|}{\textbf{100}} &
  \textbf{100} \\ \cline{2-17} 
\multicolumn{1}{|c|}{} &
  Warehouse &
  \multicolumn{1}{c|}{\textbf{100}} &
  \multicolumn{1}{c|}{92} &
  \multicolumn{1}{c|}{60} &
  \multicolumn{1}{c|}{\textbf{100}} &
  \multicolumn{1}{c|}{\textbf{100}} &
  \multicolumn{1}{c|}{92} &
  \multicolumn{1}{c|}{72} &
  \multicolumn{1}{c|}{26} &
  \multicolumn{1}{c|}{4} &
  \multicolumn{1}{c|}{76} &
  \multicolumn{1}{c|}{24} &
  \multicolumn{1}{c|}{12} &
  \multicolumn{1}{c|}{\textbf{100}} &
  \multicolumn{1}{c|}{\textbf{100}} &
  \textbf{100} \\ \cline{2-17} 
\multicolumn{1}{|c|}{} &
  Clutter &
  \multicolumn{1}{c|}{\textbf{100}} &
  \multicolumn{1}{c|}{96} &
  \multicolumn{1}{c|}{86} &
  \multicolumn{1}{c|}{\textbf{100}} &
  \multicolumn{1}{c|}{\textbf{100}} &
  \multicolumn{1}{c|}{98} &
  \multicolumn{1}{c|}{76} &
  \multicolumn{1}{c|}{28} &
  \multicolumn{1}{c|}{10} &
  \multicolumn{1}{c|}{88} &
  \multicolumn{1}{c|}{42} &
  \multicolumn{1}{c|}{18} &
  \multicolumn{1}{c|}{\textbf{100}} &
  \multicolumn{1}{c|}{\textbf{100}} &
  \textbf{100} \\ \hline
\multicolumn{1}{|c|}{\multirow{3}{*}{\begin{tabular}[c]{@{}c@{}}Comp. Time\\ (sec)\end{tabular}}} &
  Shopping mall &
  \multicolumn{1}{c|}{1.18} &
  \multicolumn{1}{c|}{7.62} &
  \multicolumn{1}{c|}{20.88} &
  \multicolumn{1}{c|}{0.08} &
  \multicolumn{1}{c|}{3.97} &
  \multicolumn{1}{c|}{14.91} &
  \multicolumn{1}{c|}{27.07} &
  \multicolumn{1}{c|}{35.36} &
  \multicolumn{1}{c|}{10.55} &
  \multicolumn{1}{c|}{10.31} &
  \multicolumn{1}{c|}{17.33} &
  \multicolumn{1}{c|}{38.69} &
  \multicolumn{1}{c|}{\textbf{0.02}} &
  \multicolumn{1}{c|}{\textbf{0.03}} &
  \textbf{0.04} \\ \cline{2-17} 
\multicolumn{1}{|c|}{} &
  Warehouse &
  \multicolumn{1}{c|}{0.07} &
  \multicolumn{1}{c|}{2.79} &
  \multicolumn{1}{c|}{12.64} &
  \multicolumn{1}{c|}{0.04} &
  \multicolumn{1}{c|}{0.19} &
  \multicolumn{1}{c|}{22.92} &
  \multicolumn{1}{c|}{11.52} &
  \multicolumn{1}{c|}{33.35} &
  \multicolumn{1}{c|}{41.95} &
  \multicolumn{1}{c|}{14.25} &
  \multicolumn{1}{c|}{41.11} &
  \multicolumn{1}{c|}{33.38} &
  \multicolumn{1}{c|}{\textbf{0.03}} &
  \multicolumn{1}{c|}{\textbf{0.05}} &
  \textbf{0.08} \\ \cline{2-17} 
\multicolumn{1}{|c|}{} &
  Clutter &
  \multicolumn{1}{c|}{0.06} &
  \multicolumn{1}{c|}{2.01} &
  \multicolumn{1}{c|}{12.37} &
  \multicolumn{1}{c|}{0.08} &
  \multicolumn{1}{c|}{1.27} &
  \multicolumn{1}{c|}{0.76} &
  \multicolumn{1}{c|}{17.54} &
  \multicolumn{1}{c|}{59.71} &
  \multicolumn{1}{c|}{64.84} &
  \multicolumn{1}{c|}{8.57} &
  \multicolumn{1}{c|}{36.30} &
  \multicolumn{1}{c|}{43.41} &
  \multicolumn{1}{c|}{\textbf{0.05}} &
  \multicolumn{1}{c|}{\textbf{0.10}} &
  \textbf{0.15} \\ \hline
\multicolumn{1}{|c|}{\multirow{3}{*}{\begin{tabular}[c]{@{}c@{}}Success rate\\ (\%)\end{tabular}}} &
  Shopping mall &
  \multicolumn{1}{c|}{100} &
  \multicolumn{1}{c|}{100} &
  \multicolumn{1}{c|}{100} &
  \multicolumn{1}{c|}{100} &
  \multicolumn{1}{c|}{100} &
  \multicolumn{1}{c|}{100} &
  \multicolumn{1}{c|}{100} &
  \multicolumn{1}{c|}{100} &
  \multicolumn{1}{c|}{100} &
  \multicolumn{1}{c|}{100} &
  \multicolumn{1}{c|}{100} &
  \multicolumn{1}{c|}{100} &
  \multicolumn{1}{c|}{100} &
  \multicolumn{1}{c|}{100} &
  96 \\ \cline{2-17} 
\multicolumn{1}{|c|}{} &
  Warehouse &
  \multicolumn{1}{c|}{100} &
  \multicolumn{1}{c|}{100} &
  \multicolumn{1}{c|}{100} &
  \multicolumn{1}{c|}{100} &
  \multicolumn{1}{c|}{100} &
  \multicolumn{1}{c|}{100} &
  \multicolumn{1}{c|}{100} &
  \multicolumn{1}{c|}{100} &
  \multicolumn{1}{c|}{100} &
  \multicolumn{1}{c|}{100} &
  \multicolumn{1}{c|}{100} &
  \multicolumn{1}{c|}{100} &
  \multicolumn{1}{c|}{100} &
  \multicolumn{1}{c|}{96} &
  96 \\ \cline{2-17} 
\multicolumn{1}{|c|}{} &
  Clutter &
  \multicolumn{1}{c|}{100} &
  \multicolumn{1}{c|}{100} &
  \multicolumn{1}{c|}{100} &
  \multicolumn{1}{c|}{100} &
  \multicolumn{1}{c|}{100} &
  \multicolumn{1}{c|}{100} &
  \multicolumn{1}{c|}{100} &
  \multicolumn{1}{c|}{100} &
  \multicolumn{1}{c|}{100} &
  \multicolumn{1}{c|}{100} &
  \multicolumn{1}{c|}{100} &
  \multicolumn{1}{c|}{100} &
  \multicolumn{1}{c|}{98} &
  \multicolumn{1}{c|}{96} &
  96 \\ \hline
\multicolumn{1}{|c|}{\multirow{3}{*}{\begin{tabular}[c]{@{}c@{}}Makespan\\ (sec)\end{tabular}}} &
  Shopping mall &
  \multicolumn{1}{c|}{4.76} &
  \multicolumn{1}{c|}{4.43} &
  \multicolumn{1}{c|}{4.22} &
  \multicolumn{1}{c|}{4.75} &
  \multicolumn{1}{c|}{4.92} &
  \multicolumn{1}{c|}{4.94} &
  \multicolumn{1}{c|}{6.07} &
  \multicolumn{1}{c|}{6.51} &
  \multicolumn{1}{c|}{4.81} &
  \multicolumn{1}{c|}{6.68} &
  \multicolumn{1}{c|}{7.03} &
  \multicolumn{1}{c|}{6.26} &
  \multicolumn{1}{c|}{\textbf{4.05}} &
  \multicolumn{1}{c|}{\textbf{3.73}} &
  \textbf{3.56} \\ \cline{2-17} 
\multicolumn{1}{|c|}{} &
  Warehouse &
  \multicolumn{1}{c|}{4.45} &
  \multicolumn{1}{c|}{4.19} &
  \multicolumn{1}{c|}{4.09} &
  \multicolumn{1}{c|}{4.57} &
  \multicolumn{1}{c|}{4.40} &
  \multicolumn{1}{c|}{5.24} &
  \multicolumn{1}{c|}{6.33} &
  \multicolumn{1}{c|}{5.88} &
  \multicolumn{1}{c|}{5.11} &
  \multicolumn{1}{c|}{6.12} &
  \multicolumn{1}{c|}{6.23} &
  \multicolumn{1}{c|}{4.74} &
  \multicolumn{1}{c|}{\textbf{4.15}} &
  \multicolumn{1}{c|}{\textbf{3.96}} &
  \textbf{3.67} \\ \cline{2-17} 
\multicolumn{1}{|c|}{} &
  Clutter &
  \multicolumn{1}{c|}{5.13} &
  \multicolumn{1}{c|}{4.97} &
  \multicolumn{1}{c|}{4.38} &
  \multicolumn{1}{c|}{5.12} &
  \multicolumn{1}{c|}{5.19} &
  \multicolumn{1}{c|}{4.90} &
  \multicolumn{1}{c|}{7.41} &
  \multicolumn{1}{c|}{7.02} &
  \multicolumn{1}{c|}{5.02} &
  \multicolumn{1}{c|}{7.28} &
  \multicolumn{1}{c|}{7.48} &
  \multicolumn{1}{c|}{6.79} &
  \multicolumn{1}{c|}{\textbf{4.31}} &
  \multicolumn{1}{c|}{\textbf{4.01}} &
  \textbf{3.78} \\ \hline
\multicolumn{1}{|c|}{\multirow{3}{*}{\begin{tabular}[c]{@{}c@{}}SoC\\ (sec)\end{tabular}}} &
  Shopping mall &
  \multicolumn{1}{c|}{20.84} &
  \multicolumn{1}{c|}{32.55} &
  \multicolumn{1}{c|}{40.65} &
  \multicolumn{1}{c|}{20.91} &
  \multicolumn{1}{c|}{33.07} &
  \multicolumn{1}{c|}{44.03} &
  \multicolumn{1}{c|}{31.58} &
  \multicolumn{1}{c|}{58.55} &
  \multicolumn{1}{c|}{75.71} &
  \multicolumn{1}{c|}{34.57} &
  \multicolumn{1}{c|}{62.42} &
  \multicolumn{1}{c|}{81.38} &
  \multicolumn{1}{c|}{\textbf{19.07}} &
  \multicolumn{1}{c|}{\textbf{29.86}} &
  \textbf{38.97} \\ \cline{2-17} 
\multicolumn{1}{|c|}{} &
  Warehouse &
  \multicolumn{1}{c|}{20.68} &
  \multicolumn{1}{c|}{32.77} &
  \multicolumn{1}{c|}{42.06} &
  \multicolumn{1}{c|}{20.76} &
  \multicolumn{1}{c|}{34.12} &
  \multicolumn{1}{c|}{45.33} &
  \multicolumn{1}{c|}{33.33} &
  \multicolumn{1}{c|}{52.29} &
  \multicolumn{1}{c|}{72.78} &
  \multicolumn{1}{c|}{33.10} &
  \multicolumn{1}{c|}{54.62} &
  \multicolumn{1}{c|}{71.83} &
  \multicolumn{1}{c|}{\textbf{20.18}} &
  \multicolumn{1}{c|}{\textbf{32.33}} &
  \textbf{41.33} \\ \cline{2-17} 
\multicolumn{1}{|c|}{} &
  Clutter &
  \multicolumn{1}{c|}{23.77} &
  \multicolumn{1}{c|}{36.72} &
  \multicolumn{1}{c|}{45.11} &
  \multicolumn{1}{c|}{23.76} &
  \multicolumn{1}{c|}{37.17} &
  \multicolumn{1}{c|}{47.24} &
  \multicolumn{1}{c|}{36.68} &
  \multicolumn{1}{c|}{56.95} &
  \multicolumn{1}{c|}{66.91} &
  \multicolumn{1}{c|}{36.97} &
  \multicolumn{1}{c|}{64.44} &
  \multicolumn{1}{c|}{85.35} &
  \multicolumn{1}{c|}{\textbf{20.50}} &
  \multicolumn{1}{c|}{\textbf{31.69}} &
  \textbf{40.70} \\ \hline \hline
\multicolumn{2}{|c|}{Distribution} &
  \multicolumn{15}{c|}{Separate} \\ \hline
\multicolumn{2}{|c|}{Method} &
  \multicolumn{3}{c|}{CBS-TA} &
  \multicolumn{3}{c|}{ECBS-TA} &
  \multicolumn{3}{c|}{TA-CCBS} &
  \multicolumn{3}{c|}{TA-CECBS} &
  \multicolumn{3}{c|}{MRTA-RM} \\ \hline
\multicolumn{2}{|c|}{\#Robots} &
  \multicolumn{1}{c|}{10} &
  \multicolumn{1}{c|}{20} &
  \multicolumn{1}{c|}{30} &
  \multicolumn{1}{c|}{10} &
  \multicolumn{1}{c|}{20} &
  \multicolumn{1}{c|}{30} &
  \multicolumn{1}{c|}{10} &
  \multicolumn{1}{c|}{20} &
  \multicolumn{1}{c|}{30} &
  \multicolumn{1}{c|}{10} &
  \multicolumn{1}{c|}{20} &
  \multicolumn{1}{c|}{30} &
  \multicolumn{1}{c|}{10} &
  \multicolumn{1}{c|}{20} &
  30 \\ \hline
\multicolumn{1}{|c|}{\multirow{3}{*}{\begin{tabular}[c]{@{}c@{}}Comp.\\ Success rate\\ (\%)\end{tabular}}} &
  Shopping mall &
  \multicolumn{1}{c|}{72} &
  \multicolumn{1}{c|}{0} &
  \multicolumn{1}{c|}{0} &
  \multicolumn{1}{c|}{86} &
  \multicolumn{1}{c|}{12} &
  \multicolumn{1}{c|}{0} &
  \multicolumn{1}{c|}{4} &
  \multicolumn{1}{c|}{0} &
  \multicolumn{1}{c|}{0} &
  \multicolumn{1}{c|}{26} &
  \multicolumn{1}{c|}{0} &
  \multicolumn{1}{c|}{0} &
  \multicolumn{1}{c|}{\textbf{100}} &
  \multicolumn{1}{c|}{\textbf{100}} &
  \textbf{100} \\ \cline{2-17} 
\multicolumn{1}{|c|}{} &
  Warehouse &
  \multicolumn{1}{c|}{96} &
  \multicolumn{1}{c|}{26} &
  \multicolumn{1}{c|}{0} &
  \multicolumn{1}{c|}{92} &
  \multicolumn{1}{c|}{24} &
  \multicolumn{1}{c|}{0} &
  \multicolumn{1}{c|}{18} &
  \multicolumn{1}{c|}{0} &
  \multicolumn{1}{c|}{0} &
  \multicolumn{1}{c|}{50} &
  \multicolumn{1}{c|}{2} &
  \multicolumn{1}{c|}{0} &
  \multicolumn{1}{c|}{\textbf{100}} &
  \multicolumn{1}{c|}{\textbf{100}} &
  \textbf{100} \\ \cline{2-17} 
\multicolumn{1}{|c|}{} &
  Clutter &
  \multicolumn{1}{c|}{20} &
  \multicolumn{1}{c|}{0} &
  \multicolumn{1}{c|}{0} &
  \multicolumn{1}{c|}{94} &
  \multicolumn{1}{c|}{20} &
  \multicolumn{1}{c|}{0} &
  \multicolumn{1}{c|}{0} &
  \multicolumn{1}{c|}{0} &
  \multicolumn{1}{c|}{0} &
  \multicolumn{1}{c|}{0} &
  \multicolumn{1}{c|}{0} &
  \multicolumn{1}{c|}{0} &
  \multicolumn{1}{c|}{\textbf{100}} &
  \multicolumn{1}{c|}{\textbf{100}} &
  \textbf{100} \\ \hline
\multicolumn{1}{|c|}{\multirow{3}{*}{\begin{tabular}[c]{@{}c@{}}Comp. Time\\ (sec)\end{tabular}}} &
  Shopping mall &
  \multicolumn{1}{c|}{21.61} &
  \multicolumn{1}{c|}{-} &
  \multicolumn{1}{c|}{-} &
  \multicolumn{1}{c|}{10.62} &
  \multicolumn{1}{c|}{58.94} &
  \multicolumn{1}{c|}{-} &
  \multicolumn{1}{c|}{92.40} &
  \multicolumn{1}{c|}{-} &
  \multicolumn{1}{c|}{-} &
  \multicolumn{1}{c|}{78.70} &
  \multicolumn{1}{c|}{-} &
  \multicolumn{1}{c|}{-} &
  \multicolumn{1}{c|}{\textbf{0.02}} &
  \multicolumn{1}{c|}{\textbf{0.03}} &
  \textbf{0.05} \\ \cline{2-17} 
\multicolumn{1}{|c|}{} &
  Warehouse &
  \multicolumn{1}{c|}{17.30} &
  \multicolumn{1}{c|}{65.62} &
  \multicolumn{1}{c|}{-} &
  \multicolumn{1}{c|}{5.69} &
  \multicolumn{1}{c|}{57.16} &
  \multicolumn{1}{c|}{-} &
  \multicolumn{1}{c|}{38.79} &
  \multicolumn{1}{c|}{-} &
  \multicolumn{1}{c|}{-} &
  \multicolumn{1}{c|}{30.98} &
  \multicolumn{1}{c|}{78.21} &
  \multicolumn{1}{c|}{-} &
  \multicolumn{1}{c|}{\textbf{0.03}} &
  \multicolumn{1}{c|}{\textbf{0.05}} &
  \textbf{0.08} \\ \cline{2-17} 
\multicolumn{1}{|c|}{} &
  Clutter &
  \multicolumn{1}{c|}{24.06} &
  \multicolumn{1}{c|}{-} &
  \multicolumn{1}{c|}{-} &
  \multicolumn{1}{c|}{18.39} &
  \multicolumn{1}{c|}{72.32} &
  \multicolumn{1}{c|}{-} &
  \multicolumn{1}{c|}{-} &
  \multicolumn{1}{c|}{-} &
  \multicolumn{1}{c|}{-} &
  \multicolumn{1}{c|}{-} &
  \multicolumn{1}{c|}{-} &
  \multicolumn{1}{c|}{-} &
  \multicolumn{1}{c|}{\textbf{0.05}} &
  \multicolumn{1}{c|}{\textbf{0.10}} &
  \textbf{0.15} \\ \hline
\multicolumn{1}{|c|}{\multirow{3}{*}{\begin{tabular}[c]{@{}c@{}}Success rate\\ (\%)\end{tabular}}} &
  Shopping mall &
  \multicolumn{1}{c|}{100} &
  \multicolumn{1}{c|}{-} &
  \multicolumn{1}{c|}{-} &
  \multicolumn{1}{c|}{100} &
  \multicolumn{1}{c|}{100} &
  \multicolumn{1}{c|}{-} &
  \multicolumn{1}{c|}{100} &
  \multicolumn{1}{c|}{-} &
  \multicolumn{1}{c|}{-} &
  \multicolumn{1}{c|}{100} &
  \multicolumn{1}{c|}{-} &
  \multicolumn{1}{c|}{-} &
  \multicolumn{1}{c|}{98} &
  \multicolumn{1}{c|}{70} &
  66 \\ \cline{2-17} 
\multicolumn{1}{|c|}{} &
  Warehouse &
  \multicolumn{1}{c|}{100} &
  \multicolumn{1}{c|}{100} &
  \multicolumn{1}{c|}{-} &
  \multicolumn{1}{c|}{100} &
  \multicolumn{1}{c|}{100} &
  \multicolumn{1}{c|}{-} &
  \multicolumn{1}{c|}{100} &
  \multicolumn{1}{c|}{-} &
  \multicolumn{1}{c|}{-} &
  \multicolumn{1}{c|}{100} &
  \multicolumn{1}{c|}{100} &
  \multicolumn{1}{c|}{-} &
  \multicolumn{1}{c|}{96} &
  \multicolumn{1}{c|}{94} &
  80 \\ \cline{2-17} 
\multicolumn{1}{|c|}{} &
  Clutter &
  \multicolumn{1}{c|}{100} &
  \multicolumn{1}{c|}{-} &
  \multicolumn{1}{c|}{-} &
  \multicolumn{1}{c|}{100} &
  \multicolumn{1}{c|}{100} &
  \multicolumn{1}{c|}{-} &
  \multicolumn{1}{c|}{-} &
  \multicolumn{1}{c|}{-} &
  \multicolumn{1}{c|}{-} &
  \multicolumn{1}{c|}{-} &
  \multicolumn{1}{c|}{-} &
  \multicolumn{1}{c|}{-} &
  \multicolumn{1}{c|}{96} &
  \multicolumn{1}{c|}{84} &
  58 \\ \hline
\multicolumn{1}{|c|}{\multirow{3}{*}{\begin{tabular}[c]{@{}c@{}}Makespan\\ (sec)\end{tabular}}} &
  Shopping mall &
  \multicolumn{1}{c|}{7.84} &
  \multicolumn{1}{c|}{-} &
  \multicolumn{1}{c|}{-} &
  \multicolumn{1}{c|}{8.34} &
  \multicolumn{1}{c|}{9.21} &
  \multicolumn{1}{c|}{-} &
  \multicolumn{1}{c|}{14.09} &
  \multicolumn{1}{c|}{-} &
  \multicolumn{1}{c|}{-} &
  \multicolumn{1}{c|}{14.40} &
  \multicolumn{1}{c|}{-} &
  \multicolumn{1}{c|}{-} &
  \multicolumn{1}{c|}{\textbf{7.15}} &
  \multicolumn{1}{c|}{\textbf{8.64}} &
  \textbf{9.89} \\ \cline{2-17} 
\multicolumn{1}{|c|}{} &
  Warehouse &
  \multicolumn{1}{c|}{7.38} &
  \multicolumn{1}{c|}{6.99} &
  \multicolumn{1}{c|}{-} &
  \multicolumn{1}{c|}{7.39} &
  \multicolumn{1}{c|}{8.84} &
  \multicolumn{1}{c|}{-} &
  \multicolumn{1}{c|}{10.72} &
  \multicolumn{1}{c|}{-} &
  \multicolumn{1}{c|}{-} &
  \multicolumn{1}{c|}{10.83} &
  \multicolumn{1}{c|}{13.59} &
  \multicolumn{1}{c|}{-} &
  \multicolumn{1}{c|}{\textbf{6.98}} &
  \multicolumn{1}{c|}{\textbf{6.77}} &
  \textbf{7.94} \\ \cline{2-17} 
\multicolumn{1}{|c|}{} &
  Clutter &
  \multicolumn{1}{c|}{11.10} &
  \multicolumn{1}{c|}{-} &
  \multicolumn{1}{c|}{-} &
  \multicolumn{1}{c|}{11.48} &
  \multicolumn{1}{c|}{14.51} &
  \multicolumn{1}{c|}{-} &
  \multicolumn{1}{c|}{-} &
  \multicolumn{1}{c|}{-} &
  \multicolumn{1}{c|}{-} &
  \multicolumn{1}{c|}{-} &
  \multicolumn{1}{c|}{-} &
  \multicolumn{1}{c|}{-} &
  \multicolumn{1}{c|}{\textbf{8.60}} &
  \multicolumn{1}{c|}{\textbf{10.05}} &
  \textbf{11.22} \\ \hline
\multicolumn{1}{|c|}{\multirow{3}{*}{\begin{tabular}[c]{@{}c@{}}SoC\\ (sec)\end{tabular}}} &
  Shopping mall &
  \multicolumn{1}{c|}{61.67} &
  \multicolumn{1}{c|}{-} &
  \multicolumn{1}{c|}{-} &
  \multicolumn{1}{c|}{63.12} &
  \multicolumn{1}{c|}{138.69} &
  \multicolumn{1}{c|}{-} &
  \multicolumn{1}{c|}{110.78} &
  \multicolumn{1}{c|}{-} &
  \multicolumn{1}{c|}{-} &
  \multicolumn{1}{c|}{114.79} &
  \multicolumn{1}{c|}{-} &
  \multicolumn{1}{c|}{-} &
  \multicolumn{1}{c|}{\textbf{56.99}} &
  \multicolumn{1}{c|}{\textbf{120.72}} &
  \textbf{191.13} \\ \cline{2-17} 
\multicolumn{1}{|c|}{} &
  Warehouse &
  \multicolumn{1}{c|}{52.86} &
  \multicolumn{1}{c|}{97.33} &
  \multicolumn{1}{c|}{-} &
  \multicolumn{1}{c|}{53.11} &
  \multicolumn{1}{c|}{105.11} &
  \multicolumn{1}{c|}{-} &
  \multicolumn{1}{c|}{84.07} &
  \multicolumn{1}{c|}{-} &
  \multicolumn{1}{c|}{-} &
  \multicolumn{1}{c|}{84.51} &
  \multicolumn{1}{c|}{202.98} &
  \multicolumn{1}{c|}{-} &
  \multicolumn{1}{c|}{\textbf{49.97}} &
  \multicolumn{1}{c|}{\textbf{93.54}} &
  \textbf{142.86} \\ \cline{2-17} 
\multicolumn{1}{|c|}{} &
  Clutter &
  \multicolumn{1}{c|}{90.99} &
  \multicolumn{1}{c|}{-} &
  \multicolumn{1}{c|}{-} &
  \multicolumn{1}{c|}{94.14} &
  \multicolumn{1}{c|}{228.42} &
  \multicolumn{1}{c|}{-} &
  \multicolumn{1}{c|}{-} &
  \multicolumn{1}{c|}{-} &
  \multicolumn{1}{c|}{-} &
  \multicolumn{1}{c|}{-} &
  \multicolumn{1}{c|}{-} &
  \multicolumn{1}{c|}{-} &
  \multicolumn{1}{c|}{\textbf{72.09}} &
  \multicolumn{1}{c|}{\textbf{152.51}} &
  \textbf{244.98} \\ \hline
\end{tabular}%
}
\vspace{-15pt}
\end{table*}

\subsection{Comparison with the methods resolving conflicts explicitly}
\label{sec:small}

We compare our method with task allocation algorithms using CBS such as CBS-TA and ECBS-TA reviewed in Sec.~\ref{related_work}. CBS-TA is one of the state-of-the-art (SOTA) methods for MRTA considering congestion during navigation. Along with its enhanced version ECBS-TA, CBS-TA is widely recognized for its effectiveness in discrete environments. Since they only work in discrete space, we discretize the environments shown in Fig.~\ref{fig:mini_envs} into grids where each cell is with 20$\times$20 units. We also develop TA-CCBS and TA-CECBS that solve task allocation and path planning separately. These latter two methods computes an allocation using MRTA-RM and then run continuous variants of CBS and ECBS which are Continuous CBS (CCBS) and Continuous ECBS (CECBS) provided by~\cite{yu2023accelerating}. While other methods could also be used, MRTA-RM is used for task allocation because it shows the best performance in our experiments, allowing us to maximize the effectiveness of the task assignment process before the path planning capabilities of CCBS and CECBS. In TA-CCBS and TA-CECBS, CBS and ECBS run on the probabilistic roadmap (PRM) to find paths in continuous space. We chose PRM over our GVD-based roadmap because GVD often yields a single representative path through wide corridors, which may limit the flexibility needed for conflict resolution in CBS-style methods. PRM provides a richer set of alternative paths, allowing more effective re-planning during CBS execution. These methods first compute a task allocation using MRTA-RM and then generate collision-free paths to execute the allocation. We increase the numbers of robots and tasks $N$ from 5 to 30 at intervals of 5. We test 50 random instances for each combination of the environment, scenario, and $N$.

For each environment, a roadmap should be constructed. This roadmap is identical for all combinations of scenarios and $N$ so only needs to be constructed once for each environment. For example, we need the roadmap construction only once to test 600 instances for the shopping mall (two scenarios, six values of $N$, and 50 instances for each combination). Thus we consider this as a preliminary step so does not included in measuring the computation time. This step takes 0.39, 1.08, and 3.08 seconds for the shopping mall, warehouse, and the clutter environments, respectively.\footnote{Even if this time for roadmap generation is included, our method still significantly outperforms the compared methods in the most cases.}

In this experiment, we set the suboptimality bounds for ECBS-TA and TA-CECBS to be 1.1. To construct PRM for CCBS and CECBS, we sample 7000 points and set the number of neighbors for each point to be 15 for the PRM. 

As shown in Fig.~\ref{fig:mini-exp} and Table~\ref{tab:small_comp}, our proposed MRTA-RM successfully finds a solution in every instance within 0.2 seconds. In contrast, CBS-TA and ECBS-TA require significantly more computation time as the number of robots increases. For instance, while CBS-TA and ECBS-TA achieved at least 60\% of the success rate with 30 robots in the random scenario, they failed to find any solutions within the 5-minute time limit in all instances in the separated scenario. This discrepancy can arise from the more frequent conflicts incurred by those paths shared by multiple robots in the separated scenario.

With respect to the computation time, MRTA-RM shows a reduction of at least 19.32\% compared to CBS-TA and 29.43\% compared to ECBS-TA in the random scenario. The experiments involving TA-CCBS and TA-CECBS in continuous space show lower success rates compared to those with CBS-TA and ECBS-TA in grid space because of the larger search space in the PRM. Furthermore, TA-CCBS and TA-CECBS have higher computation times than MRTA-RM. In the shopping mall environment with the separated scenario, TA-CCBS takes an average of 16.88 seconds for five agents and 92.40 seconds for ten agents, while TA-CECBS takes 20.29 seconds for five agents and 78.70 seconds for ten agents. In contrast, MRTA-RM requires only 0.01 to 0.05 seconds, indicating a reduction in computation time of over 99\% compared to TA-CCBS and TA-CECBS.

In the dynamic simulation, since the CBS-based methods are designed to find collision-free paths, their solutions can be executed without any conflict or deadlock. Thus, the success rate is 100\% as long as they find solutions within the time limit. However, our method cannot achieve deadlock-free as our research aims to find a task allocation that is likely to prevent deadlocks but does not set out to find collision-free paths directly. Nevertheless, our method achieves high success rates exceeding 96\% in the random scenario. Additionally, in the separated scenario, the success rate ranges from 58 to 100\% while all CBS-based methods mostly fail to compute any solution within the time limit due to their limited scalability.

The results also indicate that MRTA-RM achieves a smaller makespan and SoC compared to CBS-based methods. Specifically, MRTA-RM achieves up to a 32\% reduction in makespan and a 36\% improvement in SoC compared to ECBS-TA. When compared to CBS-TA, there is one instance in the warehouse environment involving five robots where CBS-TA has a 1\% smaller makespan. However, in all other scenarios, MRTA-RM consistently achieves a smaller makespan, with a difference of up to 24\%. In terms of SoC, MRTA-RM achieves improvements of up to 21\% over CBS-TA.

These differences become even more evident in the experiments involving TA-CCBS and TA-CECBS, which are tested in continuous space using PRM. MRTA-RM demonstrates at least a 24\% reduction in makespan and a 30\% reduction in SoC compared to TA-CCBS. Against TA-CECBS, MRTA-RM achieves reductions of at least 22\% in makespan and 32\% in SoC.

The major drawback of CBS-based methods is the computation time as they have to compute paths of all robots after finding a task allocation. In addition, they are outperformed by ours in terms of other metrics such as makespan and SoC because of the design of CBS and ECBS. In the continuous variants of CBS and ECBS, PRMs are used instead of the grid cells to represents the environments more flexibly. In the grid space, every robot has the same moving distance for a movement in each time step. However, this uniform moving distance does not hold with the topological graph (e.g., PRMs) which can have different edge lengths. Thus, in each time step, some robots moving shorter edges must wait for other robots to finish their movement in that time step, which incurs inefficiency. Even with these acknowledged limitations in CBS-based methods, our method still shows high performance in both makespan and SoC, showing its robustness and efficiency in every tested environment.

Although CBS-TA is theoretically optimal for the TAPF problem, its optimality is guaranteed only with respect to the discretized grid environment. In contrast, our method operates in continuous space by assigning tasks and recommending reference waypoints composed of JC nodes, which are followed by robots using local controllers. These controllers allow smoother navigation and flexible motion adjustments, often resulting in lower actual execution costs such as makespan and sum-of-costs, especially in dense scenarios.

\begin{figure*}[h!]
    \centering
    \includegraphics[width=0.9\textwidth]{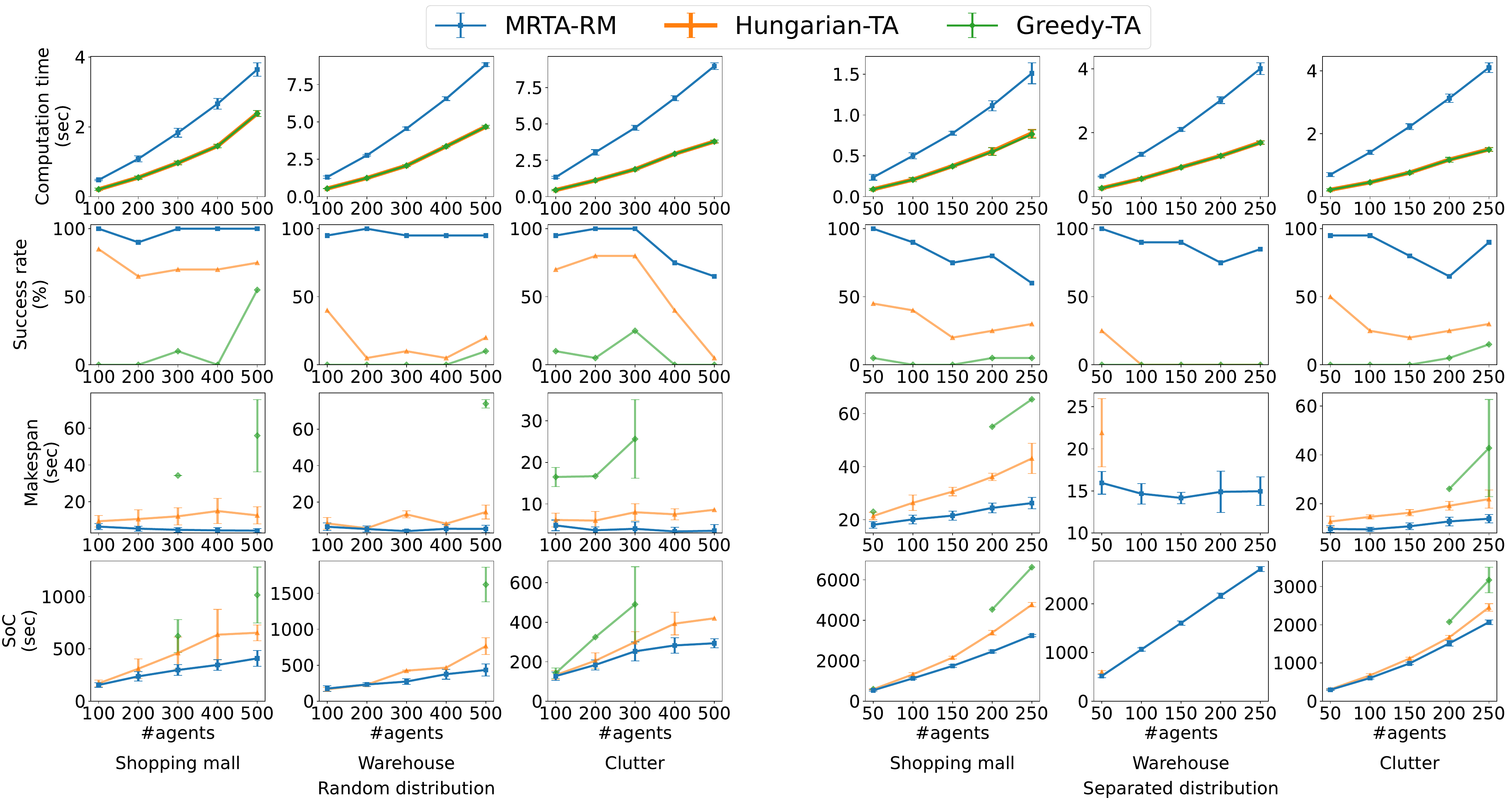}
    \vspace{-10pt}
    \caption{The comparison result with the methods that do not explicitly resolve conflicts. Our proposed method is slightly slower than others but shows significantly higher success rates in navigation.}
    \label{fig:full-exp}
\end{figure*}

\begin{table*}[h!]
\centering
\caption{The comparison result with the methods that do not explicitly resolve conflicts. Standard deviations are omitted owing to the space limit.}
\vspace{-5pt}
\label{table:full_comp}
\resizebox{\textwidth}{!}{%
\begin{tabular}{|cc||ccccccccc|ccccccccc|}
\hline
\multicolumn{2}{|c|}{Distribution} &
  \multicolumn{9}{c|}{Random} &
  \multicolumn{9}{c|}{Separate} \\ \hline
\multicolumn{2}{|c|}{Method} &
  \multicolumn{3}{c|}{Greedy-TA} &
  \multicolumn{3}{c|}{Hungarian-TA} &
  \multicolumn{3}{c|}{MRTA-RM} &
  \multicolumn{3}{c|}{Greedy-TA} &
  \multicolumn{3}{c|}{Hungarian-TA} &
  \multicolumn{3}{c|}{MRTA-RM} \\ \hline
\multicolumn{2}{|c|}{\#Robots} &
  \multicolumn{1}{c|}{100} &
  \multicolumn{1}{c|}{300} &
  \multicolumn{1}{c|}{500} &
  \multicolumn{1}{c|}{100} &
  \multicolumn{1}{c|}{300} &
  \multicolumn{1}{c|}{500} &
  \multicolumn{1}{c|}{100} &
  \multicolumn{1}{c|}{300} &
  500 &
  \multicolumn{1}{c|}{50} &
  \multicolumn{1}{c|}{150} &
  \multicolumn{1}{c|}{250} &
  \multicolumn{1}{c|}{50} &
  \multicolumn{1}{c|}{150} &
  \multicolumn{1}{c|}{250} &
  \multicolumn{1}{c|}{50} &
  \multicolumn{1}{c|}{150} &
  250 \\ \hline
\multicolumn{1}{|c|}{\multirow{3}{*}{\begin{tabular}[c]{@{}c@{}}Comp.\\ Success rate\\ (\%)\end{tabular}}} &
  Shopping mall &
  \multicolumn{1}{c|}{100} &
  \multicolumn{1}{c|}{100} &
  \multicolumn{1}{c|}{100} &
  \multicolumn{1}{c|}{100} &
  \multicolumn{1}{c|}{100} &
  \multicolumn{1}{c|}{100} &
  \multicolumn{1}{c|}{100} &
  \multicolumn{1}{c|}{100} &
  100 &
  \multicolumn{1}{c|}{100} &
  \multicolumn{1}{c|}{100} &
  \multicolumn{1}{c|}{100} &
  \multicolumn{1}{c|}{100} &
  \multicolumn{1}{c|}{100} &
  \multicolumn{1}{c|}{100} &
  \multicolumn{1}{c|}{100} &
  \multicolumn{1}{c|}{100} &
  100 \\ \cline{2-20} 
\multicolumn{1}{|c|}{} &
  Warehouse &
  \multicolumn{1}{c|}{100} &
  \multicolumn{1}{c|}{100} &
  \multicolumn{1}{c|}{100} &
  \multicolumn{1}{c|}{100} &
  \multicolumn{1}{c|}{100} &
  \multicolumn{1}{c|}{100} &
  \multicolumn{1}{c|}{100} &
  \multicolumn{1}{c|}{100} &
  100 &
  \multicolumn{1}{c|}{100} &
  \multicolumn{1}{c|}{100} &
  \multicolumn{1}{c|}{100} &
  \multicolumn{1}{c|}{100} &
  \multicolumn{1}{c|}{100} &
  \multicolumn{1}{c|}{100} &
  \multicolumn{1}{c|}{100} &
  \multicolumn{1}{c|}{100} &
  100 \\ \cline{2-20} 
\multicolumn{1}{|c|}{} &
  Clutter &
  \multicolumn{1}{c|}{100} &
  \multicolumn{1}{c|}{100} &
  \multicolumn{1}{c|}{100} &
  \multicolumn{1}{c|}{100} &
  \multicolumn{1}{c|}{100} &
  \multicolumn{1}{c|}{100} &
  \multicolumn{1}{c|}{100} &
  \multicolumn{1}{c|}{100} &
  100 &
  \multicolumn{1}{c|}{100} &
  \multicolumn{1}{c|}{100} &
  \multicolumn{1}{c|}{100} &
  \multicolumn{1}{c|}{100} &
  \multicolumn{1}{c|}{100} &
  \multicolumn{1}{c|}{100} &
  \multicolumn{1}{c|}{100} &
  \multicolumn{1}{c|}{100} &
  100 \\ \hline
\multicolumn{1}{|c|}{\multirow{3}{*}{\begin{tabular}[c]{@{}c@{}}Comp. Time\\ (sec)\end{tabular}}} &
  Shopping mall &
  \multicolumn{1}{c|}{\textbf{0.21}} &
  \multicolumn{1}{c|}{\textbf{0.97}} &
  \multicolumn{1}{c|}{\textbf{2.38}} &
  \multicolumn{1}{c|}{\textbf{0.21}} &
  \multicolumn{1}{c|}{\textbf{0.97}} &
  \multicolumn{1}{c|}{\textbf{2.38}} &
  \multicolumn{1}{c|}{0.48} &
  \multicolumn{1}{c|}{1.83} &
  3.64 &
  \multicolumn{1}{c|}{\textbf{0.09}} &
  \multicolumn{1}{c|}{\textbf{0.37}} &
  \multicolumn{1}{c|}{0.77} &
  \multicolumn{1}{c|}{\textbf{0.09}} &
  \multicolumn{1}{c|}{\textbf{0.37}} &
  \multicolumn{1}{c|}{\textbf{0.76}} &
  \multicolumn{1}{c|}{0.24} &
  \multicolumn{1}{c|}{0.78} &
  1.51 \\ \cline{2-20} 
\multicolumn{1}{|c|}{} &
  Warehouse &
  \multicolumn{1}{c|}{\textbf{0.54}} &
  \multicolumn{1}{c|}{\textbf{2.07}} &
  \multicolumn{1}{c|}{\textbf{4.67}} &
  \multicolumn{1}{c|}{\textbf{0.54}} &
  \multicolumn{1}{c|}{\textbf{2.07}} &
  \multicolumn{1}{c|}{\textbf{4.67}} &
  \multicolumn{1}{c|}{1.30} &
  \multicolumn{1}{c|}{4.55} &
  8.82 &
  \multicolumn{1}{c|}{\textbf{0.26}} &
  \multicolumn{1}{c|}{\textbf{0.92}} &
  \multicolumn{1}{c|}{\textbf{1.69}} &
  \multicolumn{1}{c|}{\textbf{0.26}} &
  \multicolumn{1}{c|}{\textbf{0.92}} &
  \multicolumn{1}{c|}{\textbf{1.69}} &
  \multicolumn{1}{c|}{0.63} &
  \multicolumn{1}{c|}{2.10} &
  4.01 \\ \cline{2-20} 
\multicolumn{1}{|c|}{} &
  Clutter &
  \multicolumn{1}{c|}{\textbf{0.45}} &
  \multicolumn{1}{c|}{\textbf{1.88}} &
  \multicolumn{1}{c|}{\textbf{3.79}} &
  \multicolumn{1}{c|}{\textbf{0.45}} &
  \multicolumn{1}{c|}{\textbf{1.88}} &
  \multicolumn{1}{c|}{\textbf{3.79}} &
  \multicolumn{1}{c|}{1.34} &
  \multicolumn{1}{c|}{4.74} &
  8.99 &
  \multicolumn{1}{c|}{\textbf{0.22}} &
  \multicolumn{1}{c|}{\textbf{0.76}} &
  \multicolumn{1}{c|}{\textbf{1.49}} &
  \multicolumn{1}{c|}{\textbf{0.22}} &
  \multicolumn{1}{c|}{\textbf{0.76}} &
  \multicolumn{1}{c|}{1.50} &
  \multicolumn{1}{c|}{0.70} &
  \multicolumn{1}{c|}{2.23} &
  4.10 \\ \hline
\multicolumn{1}{|c|}{\multirow{3}{*}{\begin{tabular}[c]{@{}c@{}}Success rate\\ (\%)\end{tabular}}} &
  Shopping mall &
  \multicolumn{1}{c|}{0} &
  \multicolumn{1}{c|}{10} &
  \multicolumn{1}{c|}{55} &
  \multicolumn{1}{c|}{85} &
  \multicolumn{1}{c|}{70} &
  \multicolumn{1}{c|}{75} &
  \multicolumn{1}{c|}{\textbf{100}} &
  \multicolumn{1}{c|}{\textbf{100}} &
  \textbf{100} &
  \multicolumn{1}{c|}{5} &
  \multicolumn{1}{c|}{0} &
  \multicolumn{1}{c|}{5} &
  \multicolumn{1}{c|}{45} &
  \multicolumn{1}{c|}{20} &
  \multicolumn{1}{c|}{30} &
  \multicolumn{1}{c|}{\textbf{100}} &
  \multicolumn{1}{c|}{\textbf{75}} &
  \textbf{60} \\ \cline{2-20} 
\multicolumn{1}{|c|}{} &
  Warehouse &
  \multicolumn{1}{c|}{0} &
  \multicolumn{1}{c|}{0} &
  \multicolumn{1}{c|}{10} &
  \multicolumn{1}{c|}{40} &
  \multicolumn{1}{c|}{10} &
  \multicolumn{1}{c|}{20} &
  \multicolumn{1}{c|}{\textbf{95}} &
  \multicolumn{1}{c|}{\textbf{95}} &
  \textbf{95} &
  \multicolumn{1}{c|}{0} &
  \multicolumn{1}{c|}{0} &
  \multicolumn{1}{c|}{0} &
  \multicolumn{1}{c|}{25} &
  \multicolumn{1}{c|}{0} &
  \multicolumn{1}{c|}{0} &
  \multicolumn{1}{c|}{\textbf{100}} &
  \multicolumn{1}{c|}{\textbf{90}} &
  \textbf{85} \\ \cline{2-20} 
\multicolumn{1}{|c|}{} &
  Clutter &
  \multicolumn{1}{c|}{10} &
  \multicolumn{1}{c|}{25} &
  \multicolumn{1}{c|}{0} &
  \multicolumn{1}{c|}{70} &
  \multicolumn{1}{c|}{80} &
  \multicolumn{1}{c|}{5} &
  \multicolumn{1}{c|}{\textbf{95}} &
  \multicolumn{1}{c|}{\textbf{100}} &
  \textbf{65} &
  \multicolumn{1}{c|}{0} &
  \multicolumn{1}{c|}{0} &
  \multicolumn{1}{c|}{15} &
  \multicolumn{1}{c|}{50} &
  \multicolumn{1}{c|}{20} &
  \multicolumn{1}{c|}{30} &
  \multicolumn{1}{c|}{\textbf{95}} &
  \multicolumn{1}{c|}{\textbf{80}} &
  \textbf{90} \\ \hline
\multicolumn{1}{|c|}{\multirow{3}{*}{\begin{tabular}[c]{@{}c@{}}Makespan\\ (sec)\end{tabular}}} &
  Shopping mall &
  \multicolumn{1}{c|}{-} &
  \multicolumn{1}{c|}{34.28} &
  \multicolumn{1}{c|}{55.96} &
  \multicolumn{1}{c|}{9.41} &
  \multicolumn{1}{c|}{12.08} &
  \multicolumn{1}{c|}{12.59} &
  \multicolumn{1}{c|}{\textbf{6.49}} &
  \multicolumn{1}{c|}{\textbf{4.72}} &
  \textbf{4.30} &
  \multicolumn{1}{c|}{23.03} &
  \multicolumn{1}{c|}{-} &
  \multicolumn{1}{c|}{65.39} &
  \multicolumn{1}{c|}{21.41} &
  \multicolumn{1}{c|}{30.59} &
  \multicolumn{1}{c|}{43.09} &
  \multicolumn{1}{c|}{\textbf{18.10}} &
  \multicolumn{1}{c|}{\textbf{21.54}} &
  \textbf{26.26} \\ \cline{2-20} 
\multicolumn{1}{|c|}{} &
  Warehouse &
  \multicolumn{1}{c|}{-} &
  \multicolumn{1}{c|}{-} &
  \multicolumn{1}{c|}{74.02} &
  \multicolumn{1}{c|}{8.27} &
  \multicolumn{1}{c|}{13.25} &
  \multicolumn{1}{c|}{14.52} &
  \multicolumn{1}{c|}{\textbf{6.43}} &
  \multicolumn{1}{c|}{\textbf{4.02}} &
  \textbf{5.26} &
  \multicolumn{1}{c|}{-} &
  \multicolumn{1}{c|}{-} &
  \multicolumn{1}{c|}{-} &
  \multicolumn{1}{c|}{21.91} &
  \multicolumn{1}{c|}{-} &
  \multicolumn{1}{c|}{-} &
  \multicolumn{1}{c|}{\textbf{15.95}} &
  \multicolumn{1}{c|}{\textbf{14.17}} &
  \textbf{14.96} \\ \cline{2-20} 
\multicolumn{1}{|c|}{} &
  Clutter &
  \multicolumn{1}{c|}{16.49} &
  \multicolumn{1}{c|}{25.62} &
  \multicolumn{1}{c|}{-} &
  \multicolumn{1}{c|}{6.14} &
  \multicolumn{1}{c|}{8.00} &
  \multicolumn{1}{c|}{8.59} &
  \multicolumn{1}{c|}{\textbf{4.83}} &
  \multicolumn{1}{c|}{\textbf{4.01}} &
  \textbf{3.53} &
  \multicolumn{1}{c|}{-} &
  \multicolumn{1}{c|}{-} &
  \multicolumn{1}{c|}{42.78} &
  \multicolumn{1}{c|}{12.68} &
  \multicolumn{1}{c|}{16.30} &
  \multicolumn{1}{c|}{21.89} &
  \multicolumn{1}{c|}{\textbf{9.65}} &
  \multicolumn{1}{c|}{\textbf{10.75}} &
  \textbf{13.85} \\ \hline
\multicolumn{1}{|c|}{\multirow{3}{*}{\begin{tabular}[c]{@{}c@{}}SoC\\ (sec)\end{tabular}}} &
  Shopping mall &
  \multicolumn{1}{c|}{-} &
  \multicolumn{1}{c|}{622.13} &
  \multicolumn{1}{c|}{1015.00} &
  \multicolumn{1}{c|}{170.61} &
  \multicolumn{1}{c|}{460.67} &
  \multicolumn{1}{c|}{654.16} &
  \multicolumn{1}{c|}{\textbf{155.35}} &
  \multicolumn{1}{c|}{\textbf{298.40}} &
  \textbf{409.12} &
  \multicolumn{1}{c|}{599.40} &
  \multicolumn{1}{c|}{-} &
  \multicolumn{1}{c|}{6621.28} &
  \multicolumn{1}{c|}{589.91} &
  \multicolumn{1}{c|}{2157.87} &
  \multicolumn{1}{c|}{4771.25} &
  \multicolumn{1}{c|}{\textbf{540.44}} &
  \multicolumn{1}{c|}{\textbf{1743.76}} &
  \textbf{3246.06} \\ \cline{2-20} 
\multicolumn{1}{|c|}{} &
  Warehouse &
  \multicolumn{1}{c|}{-} &
  \multicolumn{1}{c|}{-} &
  \multicolumn{1}{c|}{1619.99} &
  \multicolumn{1}{c|}{\textbf{166.47}} &
  \multicolumn{1}{c|}{424.36} &
  \multicolumn{1}{c|}{765.64} &
  \multicolumn{1}{c|}{175.60} &
  \multicolumn{1}{c|}{\textbf{273.07}} &
  \textbf{433.80} &
  \multicolumn{1}{c|}{-} &
  \multicolumn{1}{c|}{-} &
  \multicolumn{1}{c|}{-} &
  \multicolumn{1}{c|}{600.23} &
  \multicolumn{1}{c|}{-} &
  \multicolumn{1}{c|}{-} &
  \multicolumn{1}{c|}{\textbf{518.42}} &
  \multicolumn{1}{c|}{\textbf{1606.22}} &
  \textbf{2713.80} \\ \cline{2-20} 
\multicolumn{1}{|c|}{} &
  Clutter &
  \multicolumn{1}{c|}{142.58} &
  \multicolumn{1}{c|}{490.57} &
  \multicolumn{1}{c|}{-} &
  \multicolumn{1}{c|}{132.35} &
  \multicolumn{1}{c|}{299.98} &
  \multicolumn{1}{c|}{419.90} &
  \multicolumn{1}{c|}{\textbf{127.16}} &
  \multicolumn{1}{c|}{\textbf{252.63}} &
  \textbf{293.42} &
  \multicolumn{1}{c|}{-} &
  \multicolumn{1}{c|}{-} &
  \multicolumn{1}{c|}{3171.46} &
  \multicolumn{1}{c|}{311.01} &
  \multicolumn{1}{c|}{1119.40} &
  \multicolumn{1}{c|}{2455.43} &
  \multicolumn{1}{c|}{\textbf{298.73}} &
  \multicolumn{1}{c|}{\textbf{989.88}} &
  \textbf{2068.23} \\ \hline
\multicolumn{1}{|c|}{\multirow{3}{*}{\begin{tabular}[c]{@{}c@{}}Total time\\ (sec)\end{tabular}}} &
  Shopping mall &
  \multicolumn{1}{c|}{-} &
  \multicolumn{1}{c|}{35.25} &
  \multicolumn{1}{c|}{58.34} &
  \multicolumn{1}{c|}{9.62} &
  \multicolumn{1}{c|}{13.05} &
  \multicolumn{1}{c|}{14.97} &
  \multicolumn{1}{c|}{\textbf{6.97}} &
  \multicolumn{1}{c|}{\textbf{6.55}} &
  \textbf{7.94} &
  \multicolumn{1}{c|}{23.12} &
  \multicolumn{1}{c|}{-} &
  \multicolumn{1}{c|}{66.16} &
  \multicolumn{1}{c|}{21.50} &
  \multicolumn{1}{c|}{30.96} &
  \multicolumn{1}{c|}{43.85} &
  \multicolumn{1}{c|}{\textbf{18.34}} &
  \multicolumn{1}{c|}{\textbf{22.32}} &
  \textbf{27.77} \\ \cline{2-20} 
\multicolumn{1}{|c|}{} &
  Warehouse &
  \multicolumn{1}{c|}{-} &
  \multicolumn{1}{c|}{-} &
  \multicolumn{1}{c|}{78.69} &
  \multicolumn{1}{c|}{8.81} &
  \multicolumn{1}{c|}{15.32} &
  \multicolumn{1}{c|}{19.19} &
  \multicolumn{1}{c|}{\textbf{7.73}} &
  \multicolumn{1}{c|}{\textbf{8.57}} &
  \textbf{14.08} &
  \multicolumn{1}{c|}{-} &
  \multicolumn{1}{c|}{-} &
  \multicolumn{1}{c|}{-} &
  \multicolumn{1}{c|}{22.17} &
  \multicolumn{1}{c|}{-} &
  \multicolumn{1}{c|}{-} &
  \multicolumn{1}{c|}{\textbf{16.58}} &
  \multicolumn{1}{c|}{\textbf{16.27}} &
  \textbf{18.97} \\ \cline{2-20} 
\multicolumn{1}{|c|}{} &
  Clutter &
  \multicolumn{1}{c|}{16.94} &
  \multicolumn{1}{c|}{27.50} &
  \multicolumn{1}{c|}{-} &
  \multicolumn{1}{c|}{6.59} &
  \multicolumn{1}{c|}{9.88} &
  \multicolumn{1}{c|}{\textbf{12.38}} &
  \multicolumn{1}{c|}{\textbf{6.17}} &
  \multicolumn{1}{c|}{\textbf{8.75}} &
  12.52 &
  \multicolumn{1}{c|}{-} &
  \multicolumn{1}{c|}{-} &
  \multicolumn{1}{c|}{44.27} &
  \multicolumn{1}{c|}{12.90} &
  \multicolumn{1}{c|}{17.06} &
  \multicolumn{1}{c|}{23.39} &
  \multicolumn{1}{c|}{\textbf{10.35}} &
  \multicolumn{1}{c|}{\textbf{12.98}} &
  \textbf{17.95} \\ \hline
\end{tabular}%
}
\vspace{-7pt}
\end{table*}


\subsection{Comparisons with the methods without conflict resolution}
\label{sec:large}

To evaluate the scalability of our method in comparison to others that do not explicitly resolve conflicts, we include Greedy-TA and Hungarian-TA. They are based on the Hungarian method (Hungarian-TA) and the greedy method (Greedy-TA) which allocate tasks based on the costs from the same roadmap as ours. These methods are efficient and can handle large numbers of robots, but they do not consider conflicts during task execution, making them fundamentally different from our proposed approach. Greedy-TA assigns tasks to robots based on the lowest-cost pairing, whereas Hungarian-TA uses an optimal assignment strategy via the Hungarian method. The motivation for choosing Greedy-TA and Hungarian-TA as baselines lies in their ability to scale to extremely large instances, which is a key aspect of our evaluation. While other methods like CBS-TA, and ECBS-TA can consider both task allocation and path planning, they require significant computational overhead and are not suitable for instances involving hundreds of robots. To the best of our knowledge, there are no existing approaches that solve such large-scale MRTA problems while explicitly resolving conflicts, other than these traditional, simple methods. Thus, Hungarian-TA and Greedy-TA are included for comparison to highlight the scalability of our method and effectiveness in such challenging scenarios.

In the random scenario, $N$ grows from 100 to 500 in increments of 100. On the other hand, in the separated scenario, which is more difficult, $N$ ranges from 50 to 250 at intervals of 50. For each combination of scenario, method, and $N$, 20 instances are tested. The one-time roadmap construction takes 2.44, 17.09, and 19.16 seconds for the shopping mall, warehouse, and the clutter environments, respectively. All the compared methods including ours require the roadmaps. Again, this roadmap generation time is excluded from measuring the computation time.

The results of experiments are presented in Fig.~\ref{fig:full-exp} and Table~\ref{table:full_comp}. On average, our method takes longer to find an allocation compared to Greedy-TA and Hungarian-TA. Specifically, in the cluttered environment with 500 agents, our method requires an average of 8.99 seconds, while Hungarian-TA takes only 3.79 seconds, representing an increase of approximately 137\%. In the separated scenarios, the difference is even more pronounced, with MRTA-RM taking up to 218\% longer than Hungarian-TA in some cases. However, the difference in average computation time is at most 5.20 seconds, 
spending few more seconds to reduce huge delays in navigation is much beneficial for fleet operation as shown in the following experiment.

\begin{table*}[]
\centering
\caption{Computation time for path planning in continuous space using ST-RRT$^*$ based on our task assignments and two conventional task assignment methods using Hungarian method and greedy method. The table details the computation times required for path planning under different robot counts (100 and 200) in Warehouse and Clutter environments, with each cell showing the mean and standard deviation values, represented as mean/sd.}
\label{tab:st-rrt-time}
\resizebox{0.7\textwidth}{!}{%
\begin{tabular}{|cc||cccccc|}
\hline
\multicolumn{2}{|c|}{Distribution} &
  \multicolumn{6}{c|}{Random} \\ \hline
\multicolumn{2}{|c|}{Method} &
  \multicolumn{2}{c|}{MRTA-RM + ST-RRT$^*$} &
  \multicolumn{2}{c|}{Hungarian-TA + ST-RRT$^*$} &
  \multicolumn{2}{c|}{Greedy-TA + ST-RRT$^*$} \\ \hline
\multicolumn{2}{|c|}{\#Robots} &
  \multicolumn{1}{c|}{100} &
  \multicolumn{1}{c|}{200} &
  \multicolumn{1}{c|}{100} &
  \multicolumn{1}{c|}{200} &
  \multicolumn{1}{c|}{100} &
  200 \\ \hline
\multicolumn{1}{|c|}{\multirow{2}{*}{\begin{tabular}[c]{@{}c@{}}Comp. time (sec)\\ (mean / sd)\end{tabular}}} &
  Warehouse &
  \multicolumn{1}{c|}{26.38/18.12} &
  \multicolumn{1}{c|}{76.05/54.08} &
  \multicolumn{1}{c|}{30.30/12.96} &
  \multicolumn{1}{c|}{113.68/46.55} &
  \multicolumn{1}{c|}{53.08/37.40} &
  225.10/58.90 \\ \cline{2-8} 
\multicolumn{1}{|c|}{} &
  Clutter &
  \multicolumn{1}{c|}{24.00/7.34} &
  \multicolumn{1}{c|}{45.91/16.70} &
  \multicolumn{1}{c|}{28.07/8.64} &
  \multicolumn{1}{c|}{63.35/25.69} &
  \multicolumn{1}{c|}{46.00/38.48} &
  106.58/64.08 \\ \hline \hline
\multicolumn{2}{|c|}{Distribution} &
  \multicolumn{6}{c|}{Separate} \\ \hline
\multicolumn{2}{|c|}{Method} &
  \multicolumn{2}{c|}{MRTA-RM + ST-RRT$^*$} &
  \multicolumn{2}{c|}{Hungarian-TA + ST-RRT$^*$} &
  \multicolumn{2}{c|}{Greedy-TA + ST-RRT$^*$} \\ \hline
\multicolumn{2}{|c|}{\#Robots} &
  \multicolumn{1}{c|}{100} &
  \multicolumn{1}{c|}{200} &
  \multicolumn{1}{c|}{100} &
  \multicolumn{1}{c|}{200} &
  \multicolumn{1}{c|}{100} &
  200 \\ \hline
\multicolumn{1}{|c|}{\multirow{2}{*}{\begin{tabular}[c]{@{}c@{}}Comp. time (sec)\\ (mean / sd)\end{tabular}}} &
  Warehouse &
  \multicolumn{1}{c|}{133.72/17.48} &
  \multicolumn{1}{c|}{-} &
  \multicolumn{1}{c|}{213.49/38.54} &
  \multicolumn{1}{c|}{-} &
  \multicolumn{1}{c|}{-} &
  - \\ \cline{2-8} 
\multicolumn{1}{|c|}{} &
  Clutter &
  \multicolumn{1}{c|}{175.45/17.88} &
  \multicolumn{1}{c|}{-} &
  \multicolumn{1}{c|}{198.10/24.24} &
  \multicolumn{1}{c|}{-} &
  \multicolumn{1}{c|}{190.29/32.02} &
  - \\ \hline
\end{tabular}%
}
\vspace{-5pt}
\end{table*}

\begin{table*}[]
\centering
\caption{Performance metrics for results from ST-RRT$^*$. This table presents the makespan and SoC for paths computed using ST-RRT$^*$ under two distinct robot distributions, Random and Separate, across two environment types, Warehouse and Clutter.}

\label{tab:st-rrt-simulation}
\resizebox{0.7\textwidth}{!}{%
\begin{tabular}{|cc||cccccccc|}
\hline
\multicolumn{2}{|c|}{Distribution} &
  \multicolumn{8}{c|}{Random} \\ \hline
\multicolumn{2}{|c|}{Method} &
  \multicolumn{2}{c|}{MRTA-RM} &
  \multicolumn{2}{c|}{MRTA-RM + ST-RRT$^*$} &
  \multicolumn{2}{c|}{Hungarian-TA + ST-RRT$^*$} &
  \multicolumn{2}{c|}{Greedy-TA + ST-RRT$^*$} \\ \hline
\multicolumn{2}{|c|}{\#Robots} &
  \multicolumn{1}{c|}{100} &
  \multicolumn{1}{c|}{200} &
  \multicolumn{1}{c|}{100} &
  \multicolumn{1}{c|}{200} &
  \multicolumn{1}{c|}{100} &
  \multicolumn{1}{c|}{200} &
  \multicolumn{1}{c|}{100} &
  200 \\ \hline
\multicolumn{1}{|c|}{\multirow{2}{*}{\begin{tabular}[c]{@{}c@{}}Success rate\\ (\%)\end{tabular}}} &
  Warehouse &
  \multicolumn{1}{c|}{95} &
  \multicolumn{1}{c|}{100} &
  \multicolumn{1}{c|}{95} &
  \multicolumn{1}{c|}{50} &
  \multicolumn{1}{c|}{80} &
  \multicolumn{1}{c|}{15} &
  \multicolumn{1}{c|}{90} &
  25 \\ \cline{2-10} 
\multicolumn{1}{|c|}{} &
  Clutter &
  \multicolumn{1}{c|}{95} &
  \multicolumn{1}{c|}{100} &
  \multicolumn{1}{c|}{95} &
  \multicolumn{1}{c|}{95} &
  \multicolumn{1}{c|}{60} &
  \multicolumn{1}{c|}{0} &
  \multicolumn{1}{c|}{85} &
  70 \\ \hline
\multicolumn{1}{|c|}{\multirow{2}{*}{\begin{tabular}[c]{@{}c@{}}Makespan\\ (sec)\end{tabular}}} &
  Warehouse &
  \multicolumn{1}{c|}{6.43} &
  \multicolumn{1}{c|}{5.19} &
  \multicolumn{1}{c|}{14.95} &
  \multicolumn{1}{c|}{16.24} &
  \multicolumn{1}{c|}{21.30} &
  \multicolumn{1}{c|}{15.02} &
  \multicolumn{1}{c|}{86.15} &
  147.05 \\ \cline{2-10} 
\multicolumn{1}{|c|}{} &
  Clutter &
  \multicolumn{1}{c|}{4.83} &
  \multicolumn{1}{c|}{3.66} &
  \multicolumn{1}{c|}{9.97} &
  \multicolumn{1}{c|}{9.07} &
  \multicolumn{1}{c|}{10.36} &
  \multicolumn{1}{c|}{9.36} &
  \multicolumn{1}{c|}{55.34} &
  40.40 \\ \hline
\multicolumn{1}{|c|}{\multirow{2}{*}{\begin{tabular}[c]{@{}c@{}}SoC\\ (sec)\end{tabular}}} &
  Warehouse &
  \multicolumn{1}{c|}{175.60} &
  \multicolumn{1}{c|}{232.69} &
  \multicolumn{1}{c|}{288.12} &
  \multicolumn{1}{c|}{420.46} &
  \multicolumn{1}{c|}{316.07} &
  \multicolumn{1}{c|}{449.89} &
  \multicolumn{1}{c|}{496.97} &
  956.38 \\ \cline{2-10} 
\multicolumn{1}{|c|}{} &
  Clutter &
  \multicolumn{1}{c|}{127.16} &
  \multicolumn{1}{c|}{183.94} &
  \multicolumn{1}{c|}{158.54} &
  \multicolumn{1}{c|}{210.87} &
  \multicolumn{1}{c|}{171.43} &
  \multicolumn{1}{c|}{225.79} &
  \multicolumn{1}{c|}{283.11} &
  355.66 \\ \hline \hline
\multicolumn{2}{|c|}{Distribution} &
  \multicolumn{8}{c|}{Separate} \\ \hline
\multicolumn{2}{|c|}{Method} &
  \multicolumn{2}{c|}{MRTA-RM} &
  \multicolumn{2}{c|}{MRTA-RM + ST-RRT$^*$} &
  \multicolumn{2}{c|}{Hungarian-TA + ST-RRT$^*$} &
  \multicolumn{2}{c|}{Greedy-TA + ST-RRT$^*$} \\ \hline
\multicolumn{2}{|c|}{\#Robots} &
  \multicolumn{1}{c|}{100} &
  \multicolumn{1}{c|}{200} &
  \multicolumn{1}{c|}{100} &
  \multicolumn{1}{c|}{200} &
  \multicolumn{1}{c|}{100} &
  \multicolumn{1}{c|}{200} &
  \multicolumn{1}{c|}{100} &
  200 \\ \hline
\multicolumn{1}{|c|}{\multirow{2}{*}{\begin{tabular}[c]{@{}c@{}}Success rate\\ (\%)\end{tabular}}} &
  Warehouse &
  \multicolumn{1}{c|}{90} &
  \multicolumn{1}{c|}{75} &
  \multicolumn{1}{c|}{70} &
  \multicolumn{1}{c|}{0} &
  \multicolumn{1}{c|}{50} &
  \multicolumn{1}{c|}{0} &
  \multicolumn{1}{c|}{0} &
  0 \\ \cline{2-10} 
\multicolumn{1}{|c|}{} &
  Clutter &
  \multicolumn{1}{c|}{95} &
  \multicolumn{1}{c|}{65} &
  \multicolumn{1}{c|}{90} &
  \multicolumn{1}{c|}{0} &
  \multicolumn{1}{c|}{60} &
  \multicolumn{1}{c|}{0} &
  \multicolumn{1}{c|}{65} &
  0 \\ \hline
\multicolumn{1}{|c|}{\multirow{2}{*}{\begin{tabular}[c]{@{}c@{}}Makespan\\ (sec)\end{tabular}}} &
  Warehouse &
  \multicolumn{1}{c|}{14.67} &
  \multicolumn{1}{c|}{14.89} &
  \multicolumn{1}{c|}{55.34} &
  \multicolumn{1}{c|}{-} &
  \multicolumn{1}{c|}{140.25} &
  \multicolumn{1}{c|}{-} &
  \multicolumn{1}{c|}{-} &
  - \\ \cline{2-10} 
\multicolumn{1}{|c|}{} &
  Clutter &
  \multicolumn{1}{c|}{9.49} &
  \multicolumn{1}{c|}{12.74} &
  \multicolumn{1}{c|}{36.93} &
  \multicolumn{1}{c|}{-} &
  \multicolumn{1}{c|}{49.48} &
  \multicolumn{1}{c|}{-} &
  \multicolumn{1}{c|}{109.21} &
  - \\ \hline
\multicolumn{1}{|c|}{\multirow{2}{*}{\begin{tabular}[c]{@{}c@{}}SoC\\ (sec)\end{tabular}}} &
  Warehouse &
  \multicolumn{1}{c|}{1064.30} &
  \multicolumn{1}{c|}{2166.83} &
  \multicolumn{1}{c|}{1805.00} &
  \multicolumn{1}{c|}{-} &
  \multicolumn{1}{c|}{2422.61} &
  \multicolumn{1}{c|}{-} &
  \multicolumn{1}{c|}{-} &
  - \\ \cline{2-10} 
\multicolumn{1}{|c|}{} &
  Clutter &
  \multicolumn{1}{c|}{607.21} &
  \multicolumn{1}{c|}{1513.74} &
  \multicolumn{1}{c|}{959.10} &
  \multicolumn{1}{c|}{-} &
  \multicolumn{1}{c|}{1165.14} &
  \multicolumn{1}{c|}{-} &
  \multicolumn{1}{c|}{1321.85} &
  - \\ \hline
\end{tabular}%
}
\vspace{-15pt}
\end{table*}

In the dynamic simulation, Greedy-TA and Hungarian-TA show significant limitations due to their lack of conflict resolution mechanisms, often leading to deadlocks. As a result, they show considerably low success rates. For example, the Greedy-TA has a success rate of up to 25\% in all but one experimental environment in random scenario. This low success rate is due to the nature of the greedy method, which assigns robot and task pairs that are in close distance, blocking the path of other robots. However, in a shopping mall environment with 500 robots and tasks in a random scenario, Greedy-TA achieves a success rate of 55\%, which is attributed to the increased density of robots and tasks, which reduces the average distance between robots and tasks, thus reducing the number of times robots block the paths of each other. The Hungarian-TA method has measured higher success rates than Greedy-TA, up to 85\% in random scenarios and up to 50\% in separate scenarios. These results are because Hungarian-TA, like our proposed MRTA-RM, satisfies Property 1 in Sec.~\ref{sec:description}, which prevents head-on collisions between robots. On the other hand, MRTA-RM shows a higher success rate of at least 65\% and up to 100\% in random scenarios and at least 60\% and up to 100\% in separated scenario. In particular, MRTA-RM shows a success rate of at least 75\% in a warehouse environment with more than 100 robots and tasks separately located, while Greedy-TA and Hungarian-TA methods do not succeed even once.

For all successful instances, MRTA-RM shows smaller makespan compared to Hungarian-TA and Greedy-TA, with improvements of up to 92\% over Greedy-TA and up to 70\% over Hungarian-TA. These results also show that MRTA-RM has an advantage in terms of the total time, which is a sum of computation time and makespan, compared to conventional methods.

For environments with at least two successful instances (to compute statistics), Hungarian-TA measures an average SoC that is 5\% lower than MRTA-RM in the warehouse scenario with 100 robots and tasks in separate scenario. However, in all other cases, MRTA-RM has a lower SoC than the comparison methods, specifically, at least 11\% lower than Greedy-TA and at least 4\% lower than Hungarian-TA.

The reason MRTA-RM achieves lower makespan and SoC compared to the other methods is its strategy of implementing a robot redistribution plan to balance the component supplies. Specifically, MRTA-RM leverages robots from components that lie between oversupplied and undersupplied components, thereby effectively redistributing them.

Overall, our proposed method indicates high scalability, efficiently computing effective solutions for hundreds of robots while also reducing the likelihood of deadlocks without relying on a MAPF solver.

\subsection{Integration with MAPF solver}
\label{sec:st-rrt*}

In this experiment, we evaluate task assignment methods discussed in Sec.~\ref{sec:large} with a MAPF solver to compare the outcomes of executing the task allocation. To conduct large-scale experiments in continuous space, we use ST-RRT$^*$~\cite{st-rrt*}. The ST-RRT$^*$ is implemented in C++17, as provided by \cite{ompl}. 
The experimental results of integrating ST-RRT$^*$ with a task assignment approach are detailed in Tables~\ref{tab:st-rrt-time} and \ref{tab:st-rrt-simulation}, which present the computation times, success rates, makespan, and SoC under different robot distributions and environments. We set a 5-minute time limit for the ST-RRT$^*$ algorithm to find solutions, consistent with previous experiments.

In the random scenarios involving 200 robots, the MRTA-RM + ST-RRT$^*$ method maintains a 95\% success rate in cluttered environments while the Hungarian-TA + ST-RRT$^*$ method has a 0\% success rate and the Greedy-TA + ST-RRT$^*$ method has a 70\% success rate.

For the separated scenarios, the MRTA-RM + ST-RRT$^*$ method consistently outperforms the Hungarian-TA + ST-RRT$^*$ and Greedy-TA + ST-RRT$^*$ methods. In a warehouse environment with 100 robots and tasks, MRTA-RM achieves a 70\% success rate, substantially higher than the 50\% success rate of the Hungarian method and significantly better than the 0\% achieved by the Greedy method. In cluttered environments with the same number of robots, MRTA-RM maintains a 90\% success rate, while the Hungarian method achieves 60\%, and the Greedy method manages 65\%. With 200 robots, the ST-RRT$^*$ method cannot compute for every instance.

In the comparison of computation times for path planning using the ST-RRT$^*$ algorithm based on different task assignment strategies, our proposed MRTA-RM method consistently demonstrates superior performance. Specifically, it shows a minimum of 11.43\% faster computation time compared to the Hungarian method, and 7.80\% faster than the Greedy method. This substantial reduction in computation time highlights the efficiency of the MRTA-RM method in processing tasks faster compared to conventional methods.

In terms of the quality of the MAPF solutions, the MRTA-RM assignment strategy shows higher success rates across all cases than other task assignment methods. While the Hungarian method achieves an 8.14\% better makespan than MRTA-RM in the warehouse environment with randomly distributed 200 robots and tasks, MRTA-RM exceeds the Hungarian and other conventional task assignment strategies in all other cases by at least 3.10\% to as much as 88.95\%. Additionally, the SoC measured for MRTA-RM was consistently at least 6.54\% lower across all scenarios.

Note that the implementation of ST-RRT$^*$ in our scenarios, as discussed in \cite{st-rrt*-disadvantage}, may increase makespan and SoC due to the challenges arising from sampling in unbounded state spaces. This can lead to denser sample generation requirements and, subsequently, higher values in makespan and SoC.

These results highlight the efficiency and effectiveness of our proposed MRTA-RM task assignment strategy, particularly in its ability to handle complex scenarios with improved computational speeds and higher quality solutions.

\section{Conclusion and Future Work}
In this paper, we proposed a method for task allocation that considers the conflicts between robots while the robots execute the allocated tasks. By a roadmap-based redistribution strategy and a conflict-aware task allocation process, we achieved a balance between the supply and demand of robots across different components of the environment such that robot navigation and task execution have less conflicts and deadlocks. Our experiments with a dynamic simulation show the effectiveness of MRTA-RM even with hundreds of robots in challenging environments. In the future, we will achieve 100\% of the success rate by implementing a controller to enable the robots to follow the paths on the roadmap.

Furthermore, to extend the applicability of our method to heterogeneous robot teams with varying sizes or capabilities, we plan to investigate generating distinct GVD-based roadmaps tailored to each robot type.

\section*{Acknowledgments}
This work was supported by Korea Planning \& Evaluation Institute of Industrial Technology (KEIT) grant funded by the Korea government (MOTIE) (No. RS-2024-00444344) and the National Research Foundation of Korea (NRF) grant funded by the Korea government (MSIT) (No. 2022R1C1C1008476).%


\bibliographystyle{elsarticle-num} 
\bibliography{ref}






\end{document}